\documentclass[11pt]{article}

\usepackage[final]{EMNLP2022}

\usepackage{times}
\usepackage{latexsym}

\usepackage[T1]{fontenc}
\usepackage[english]{babel}
\usepackage[utf8]{inputenc}

\usepackage{microtype}

\usepackage{inconsolata}
\usepackage{hyperref}
\usepackage{expex}
\usepackage{url}
\usepackage{graphicx}
\usepackage{booktabs}
\usepackage{threeparttable}
\usepackage{multirow}
\usepackage{multicol}
\usepackage{makecell}
\usepackage{amsmath}
\usepackage{amssymb}
\usepackage{svg}
\usepackage[inline]{enumitem}

\usepackage{enumitem}
\usepackage{tikz}
\usepackage{pifont}%

\usepackage{subcaption}

\definecolor{Gray}{gray}{0.9}
\definecolor{cb-blue-green} {RGB}{ 0,  073,  073}
\definecolor{cb-green-sea}  {RGB}{ 0, 146, 146}
\definecolor{cb-rose}       {RGB}{255, 109, 182}
\definecolor{cb-salmon-pink}{RGB}{255, 182, 119}
\definecolor{cb-purple}     {RGB}{ 73,   0, 146}
\definecolor{cb-blue}       {RGB}{ 0, 109, 219}
\definecolor{cb-lilac}      {RGB}{182, 109, 255}
\definecolor{cb-blue-sky}   {RGB}{109, 182, 255}
\definecolor{cb-blue-light} {RGB}{182, 219, 255}
\definecolor{cb-burgundy}   {RGB}{146,   0,   0}
\definecolor{cb-brown}      {RGB}{146,  73,   0}
\definecolor{cb-clay}       {RGB}{219, 209,   0}
\definecolor{cb-green-lime} {RGB}{ 36, 255,  36}
\definecolor{cb-yellow}     {RGB}{255, 255, 109}
\definecolor{cb-grey}       {RGB}{233, 233, 233}

\newcommand{\codeemoji}[0]{\includegraphics[width=.02\textwidth]{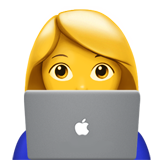}} 
\newcommand{\xmark}{\ding{55}}%
\newcommand{\cmark}{\ding{51}}%

\makeatletter
\newcommand*{\radiobutton}{%
  \@ifstar{\@radiobutton0}{\@radiobutton1}%
}
\newcommand*{\@radiobutton}[1]{%
  \begin{tikzpicture}
    \pgfmathsetlengthmacro\radius{height("X")/3}
    \draw[radius=\radius] circle;
    \ifcase#1 \fill[radius=.6*\radius] circle;\fi
  \end{tikzpicture}%
}
\makeatother

\usepackage{longtable}
\usepackage{multicol}

\title{TAPE: Assessing Few-shot Russian Language Understanding}

\author{
    ~\textbf{Ekaterina Taktasheva\textsuperscript{1,2}}\thanks{\ \ Equal contribution.},
    ~\textbf{Tatiana Shavrina\textsuperscript{1,3}$^*$},
    ~\textbf{Alena Fenogenova\textsuperscript{1}$^*$}, 
    ~\textbf{Denis Shevelev\textsuperscript{1}}, \\
    ~\textbf{Nadezhda Katricheva\textsuperscript{1}},
    ~\textbf{Maria Tikhonova\textsuperscript{1,2}}, 
    ~\textbf{Albina Akhmetgareeva\textsuperscript{1}}, \\
    ~\textbf{Oleg Zinkevich\textsuperscript{2}},
    ~\textbf{Anastasiia Bashmakova\textsuperscript{2}}, 
    ~\textbf{Svetlana Iordanskaia\textsuperscript{2}}, 
     ~\textbf{Alena Spiridonova\textsuperscript{2}}, \\
    ~\textbf{Valentina Kurenshchikova\textsuperscript{2}}, 
    ~\textbf{Ekaterina Artemova\textsuperscript{4,5}},
    ~\textbf{Vladislav Mikhailov\textsuperscript{1}} \\ 
    \textsuperscript{1}SberDevices, 
    \textsuperscript{2}HSE University,
    \textsuperscript{3}Artificial Intelligence Research Institute, \\
    \textsuperscript{4}Huawei Noah’s Ark lab,
    \textsuperscript{5}CIS LMU Munich, Germany \\
    \small{
    \textbf{Correspondence:} \href{mailto:rybolos@gmail.com}{rybolos@gmail.com}}
}

\begin{document}
\maketitle

\begin{abstract}
Recent advances in zero-shot and few-shot learning have shown promise for a scope of research and practical purposes. However, this fast-growing area lacks standardized evaluation suites for non-English languages, hindering progress outside the Anglo-centric paradigm. To address this line of research, we propose TAPE (Text Attack and Perturbation Evaluation), a novel benchmark that includes six more complex NLU tasks for Russian, covering multi-hop reasoning, ethical concepts, logic and commonsense knowledge. The TAPE's design focuses on systematic zero-shot and few-shot NLU evaluation: \emph{(i)} linguistic-oriented adversarial attacks and perturbations for analyzing robustness, and \emph{(ii)} subpopulations for nuanced interpretation. The detailed analysis of testing the autoregressive baselines indicates that simple spelling-based perturbations affect the performance the most, while paraphrasing the input has a more negligible effect. At the same time, the results demonstrate a significant gap between the neural and human baselines for most tasks. We publicly release TAPE\footnote{\href{https://tape-benchmark.com/}{tape-benchmark.com}} to foster research on robust LMs that can generalize to new tasks when little to no supervision is available.
\end{abstract}

\section{Introduction}
The ability to acquire new concepts from a few examples is central to human intelligence~\cite{tenenbaum2011grow}. Recent advances in the NLP field have fostered the development of language models (LMs;~\citealp{radford2019language,NEURIPS2020_1457c0d6}) that exhibit such generalization capacity under a wide range of few-shot learning and prompting methods~\cite{liu2021pre,beltagy-etal-2022-zero}. The community has addressed various aspects of few-shot learning, such as efficient model application~\cite{schick-schutze-2021-just}, adaptation to
unseen tasks and domains~\cite{bansal-etal-2020-learning,bansal-etal-2020-self}, and cross-lingual generalization~\cite{winata-etal-2021-language,DBLP:journals/corr/abs-2112-10668}.

The latest research has raised an essential question of standardized evaluation protocols to assess few-shot generalization from multiple perspectives. The novel tool-kits and benchmarks mainly focus on systematic evaluation design~\cite{bragg2021flex,zheng-etal-2022-fewnlu}, cross-task generalization~\cite{ye-etal-2021-crossfit,wang2022benchmarking}, and real-world scenarios~\cite{alex2021raft}. However, this rapidly developing area fails to provide similar evaluation suites for non-English languages, hindering progress outside the Anglo-centric paradigm.

\vspace{0.2em}\noindent \textbf{Motivation and Contributions.} 
In this paper, we introduce TAPE\footnote{Text Attack and Perturbation Evaluation.}, a novel benchmark for few-shot Russian language understanding evaluation. Our objective is to provide a reliable tool and methodology for nuanced assessment of zero-shot and few-shot methods for Russian. The objective is achieved through two main contributions.

\vspace{0.2em}\noindent \textbf{Contribution 1.} Our first contribution is to create six more complex question answering (QA), Winograd schema, and ethics tasks for Russian. The tasks require understanding many aspects of language, multi-hop reasoning, logic, and commonsense knowledge. 

\vspace{0.2em} \noindent The motivation behind this is that there are systems that match or outperform human baselines on most of the existing QA tasks for Russian, e.g., the ones from Russian SuperGLUE~\cite{shavrina-etal-2020-russiansuperglue}: DaNetQA~\cite{glushkova2020danetqa}, MuSeRC and RuCoS~\cite{fenogenova-etal-2020-read}. To the best of our knowledge, datasets on ethical concepts have not yet been created in Russian. To bridge this gap, we propose one of the first Russian datasets on estimating the ability of LMs to predict human ethical judgments about various text situations.

\vspace{0.2em}
\noindent \textbf{Contribution 2.} Our second contribution is to develop a framework for multifaceted zero-shot and few-shot NLU evaluation. The design includes \emph{(i)} linguistic-oriented adversarial attacks and perturbations for testing robustness, and \emph{(ii)} subpopulations for nuanced performance analysis.

\vspace{0.2em}
\noindent Here, we follow the methodological principles and recommendations by~\citet{bowman-dahl-2021-will} and \citet{bragg2021flex}, which motivate the need for systematic benchmark design and adversarially-constructed test sets.

\vspace{0.2em}
\noindent \textbf{Findings.} Our findings are summarized as five-fold: \emph{(i)} zero-shot evaluation may outperform few-shot evaluation, meaning that the autoregressive baselines fail to utilize demonstrations, \emph{(ii)} few-shot results may be unstable and sensitive to prompt changes, \emph{(iii)} \underline{negative result}: zero-shot and few-shot generation for open-domain and span selection QA tasks leads to near-zero performance, \emph{(iv)} the baselines are most vulnerable to spelling-based and emoji-based adversarial perturbations, and \emph{(v)} human annotators significantly outperform the neural baselines, indicating that there is still room for developing robust and generalizable systems.

\section{Related Work}
\noindent \textbf{Benchmark Critique.} Benchmarks such as GLUE~\cite{wang-etal-2018-glue} and SuperGLUE~\cite{wang2019superglue} have become de facto standard tools to measure progress in NLP. However, recent studies have criticized the canonical benchmarking approaches. \citet{bender2021dangers} warn performance gains are achieved at the cost of carbon footprint. \citet{elangovan-etal-2021-memorization} claim that the current benchmarks evaluate the LM's ability to memorize rather than generalize because of the significant overlap between the train and test datasets. \citet{church2022emerging} argue that benchmarks focus on relatively easy tasks instead of creating long-term challenges.  \citet{raji2021ai} raise concerns about the resource-intensive task design. In particular, benchmarks present with large-scale train datasets, which are expensive to create. This may lead to benchmark stagnation, as new tasks can not be added easily~\cite{barbosa2022mapping}. In turn, few-shot benchmarking offers a prospective avenue for LMs evaluation regarding generalization capacity, computational and resource costs.

\paragraph{Few-shot Benchmarking.} Research in few-shot benchmarking has evolved in several directions. \citet{schick-schutze-2021-just} create FewGLUE by sampling small fixed-sized training datasets from SuperGLUE. Variance w.r.t to training dataset size and sampling strategy is not reported. Later works overcome these issues by exploring evaluation strategies by $K$-fold cross-validation~\cite{perez2021true}, bagging, and multi-splits, introduced in FewNLU~\cite{zheng-etal-2022-fewnlu}. Additionally, FewNLU explores correlations between performance on development and test sets and stability w.r.t. the number of runs. CrossFit~\cite{ye-etal-2021-crossfit} studies cross-task generalization by unifying task formats and splitting tasks into training, development, and test sets. FLEX~\cite{bragg2021flex} covers the best practices and provides a unified interface for different types of transfer and varying shot sizes. Finally, to the best of our knowledge, the only non-English dataset for few-shot benchmarking is Few-CLUE in Chinese~\cite{xu2021fewclue}. TAPE is the first few-shot benchmark for Russian, which introduces variations at the data level by creating adversarial test sets.

\section{Task Formulations}
\label{sec:task_formulation}
TAPE includes six novel datasets for Russian, each requiring modeling ``intellectual abilities'' of at least two skills: logical reasoning (\S\ref{winograd}; extended Winograd schema challenge), reasoning with world knowledge (\S\ref{multiple_choice}; CheGeKa, RuOpenBookQA and RuWorldTree), multi-hop reasoning (\S\ref{multiple_choice}; MultiQ), and ethical judgments (\S\ref{ethics}; Ethics$_{1/2}$). This section describes the task formulations, general data collection stages, and dataset examples.~\autoref{app:general_statistics} provides the general dataset statistics, while Appendix~\ref{app:data_collection} includes details on dataset collection and extra validation stage via a crowd-sourcing platform Toloka\footnote{\href{https://toloka.ai/}{toloka.ai}}~\cite{toloka}.

\subsection{Logical Reasoning}
\label{winograd}
\noindent\textbf{Winograd.} The Winograd schema challenge composes tasks with syntactic ambiguity, which can be resolved with logical reasoning~\cite{levesque2012winograd}. The texts for the dataset are collected with a semi-automatic pipeline. First, lists of 11 typical grammatical structures with syntactic homonymy (mainly case) are compiled by a few authors with linguistic background (see~\autoref{app:winograd}). Queries corresponding to these constructions are submitted to the search of the Russian National Corpus\footnote{\href{https://ruscorpora.ru/old/en/index.html}{ruscorpora.ru/en}}, which includes a sub-corpus with resolved homonymy. In the resulting 2k+ sentences, homonymy is resolved automatically with UDPipe\footnote{\href{https://cran.r-project.org/web/packages/udpipe/vignettes/udpipe-annotation.html}{UDPipe package}} and then validated manually by a few authors afterward. Each sentence is split into multiple examples in the binary classification format, indicating whether the reference pronoun is dependant on the chosen candidate noun.

\begin{itemize}[noitemsep,leftmargin=1.em]
\item \textbf{Context:} \textit{``Brosh' iz Pompei, kotoraya perezhila veka.''} (A trinket from Pompeii that has survived the centuries.)
\item \textbf{Reference:} \textit{``kotoraya''} (that)%
\item \textbf{Candidate Answer:} \textit{``Brosh' ''} (A trinket) %
\item \textbf{Label:} \cmark (correct resolution)%
\end{itemize}

\subsection{Reasoning with World Knowledge}
\label{multiple_choice}
\noindent\textbf{RuOpenBookQA.} RuOpenBookQA is a QA dataset with multiple-choice elementary-level science questions, which probe understanding of 1k+ core science facts. The dataset is built with automatic translation of the original English dataset by~\citet{mihaylov-etal-2018-suit} and manual validation by a few authors.

\begin{itemize}[noitemsep,leftmargin=1.em]
\item \textbf{Question:} \textit{``Yesli chelovek idet v napravlenii, protivopolozhnom napravleniyu strelki kompasa, on idet...''} (If a person walks in the direction opposite to the compass needle, they are going...)
\item \textbf{Answers:}
    \begin {enumerate*} [label=(\Alph*\upshape)]
        \item \textit{``na zapad''} (west);
        \item \textit{``na sever''} (north);
        \item \textit{``na vostok''} (east);
        \item \underline{\textit{``na yug''} (south)}.
    \end{enumerate*}
\end{itemize}

\noindent\textbf{RuWorldTree.} The collection approach of \textbf{RuWorldTree} is similar to that of \textbf{RuOpenBookQA}, the main difference being the additional lists of facts and the logical order that is attached to the output of each answer to a question~\cite{jansen-etal-2018-worldtree}.

\begin{itemize}[noitemsep,leftmargin=1.em]
\item \textbf{Question:} \textit{``Kakoye svoystvo vody izmenitsya, kogda voda dostignet tochki zamerzaniya?''} (What property of water will change when the water reaches the freezing point?)
\item  \textbf{Answers:}
    \begin {enumerate*} [label=(\Alph*\upshape)]
        \item \textit{``tsvet''} (color);
        \item \textit{``massa''} (mass);
        \item \underline{\textit{``sostoyaniye''} (state of matter)};
        \item \textit{``ves''} (weight).
    \end{enumerate*}
\end{itemize}

\noindent\textbf{MultiQ.}
\label{multiq} Multi-hop reasoning has been one of the least explored QA directions for Russian. The task is addressed by the MuSeRC dataset~\cite{fenogenova-etal-2020-read} and only a few dozen questions in SberQUAD~\cite{Efimov_2020} and RuBQ~\cite{rybin2021rubq}. In response, we have developed a semi-automatic pipeline for multi-hop dataset generation based on Wikidata and Wikipedia.
First, we extract the triplets from Wikidata and search for their intersections. Two triplets (subject, relation, object) are needed to compose an answerable multi-hop question. For instance, the question \textit{``Na kakom kontinente nakhoditsya strana, grazhdaninom kotoroy byl Yokhannes Blok?''}
(In what continent lies the country of which Johannes Block was a citizen?) is formed by a sequence of five graph units: \textit{``Blok, Yokhannes''} (Block, Johannes), \textit{``grazhdanstvo''} (country of citizenship), \textit{``Germaniya''} (Germany), \textit{``chast' sveta''} (continent), and \textit{``Yevropa''} (Europe). Second, several hundreds of the corresponding question templates are curated by a few authors manually, which are further used to fine-tune ruT5-large\footnote{\href{https://huggingface.co/sberbank-ai/ruT5-large/tree/main}{hf.co/sberbank-ai/ruT5-large}} to generate multi-hop questions given the graph units sequences. Third, the resulting questions undergo paraphrasing~\cite{fenogenova-2021-russian} and manual validation procedure to control the quality and diversity. Finally, each question is linked to two Wikipedia paragraphs with the help of wptools\footnote{\href{https://github.com/siznax/wptools/}{github.com/siznax/wptools}}, where all graph units appear in the natural language. The task is to select the answer span using information from both paragraphs.

\begin{itemize}[noitemsep,leftmargin=1.em]
\item \textbf{Question:} \textit{``Gde nakhoditsya istok reki, pritokom kotoroy yavlyayetsya Getar?''} (Where is the source of the river, the tributary of which is the Getar?)
\item \textbf{Supporting Text:} \textit{``Getar — reka v Armenii. Beryot nachalo na territorii Kotaykskoy oblasti, protekayet cherez tsentral'nuyu chast' Yerevana i vpadayet v Razdan.''} (The Getar is a river in Armenia. [It] originates in the Kotayk region, flows through the central part of Yerevan and flows into the Hrazdan.)%
\item \textbf{Main Text:} \textit{``Razdan — reka v Armenii. Vytekayet iz ozera \colorbox{cb-salmon-pink}{Sevan} v yego severo-zapadnoy chasti, nedaleko ot goroda Sevan.''} (The Hrazdan is a river in Armenia. [It] originates at the northwest extremity of Lake Sevan, near the city of Sevan.) %
\item \textbf{Answer:} Sevan%
\end{itemize}

\begin{figure*}[!ht]
\centering
\includegraphics[width=\textwidth]{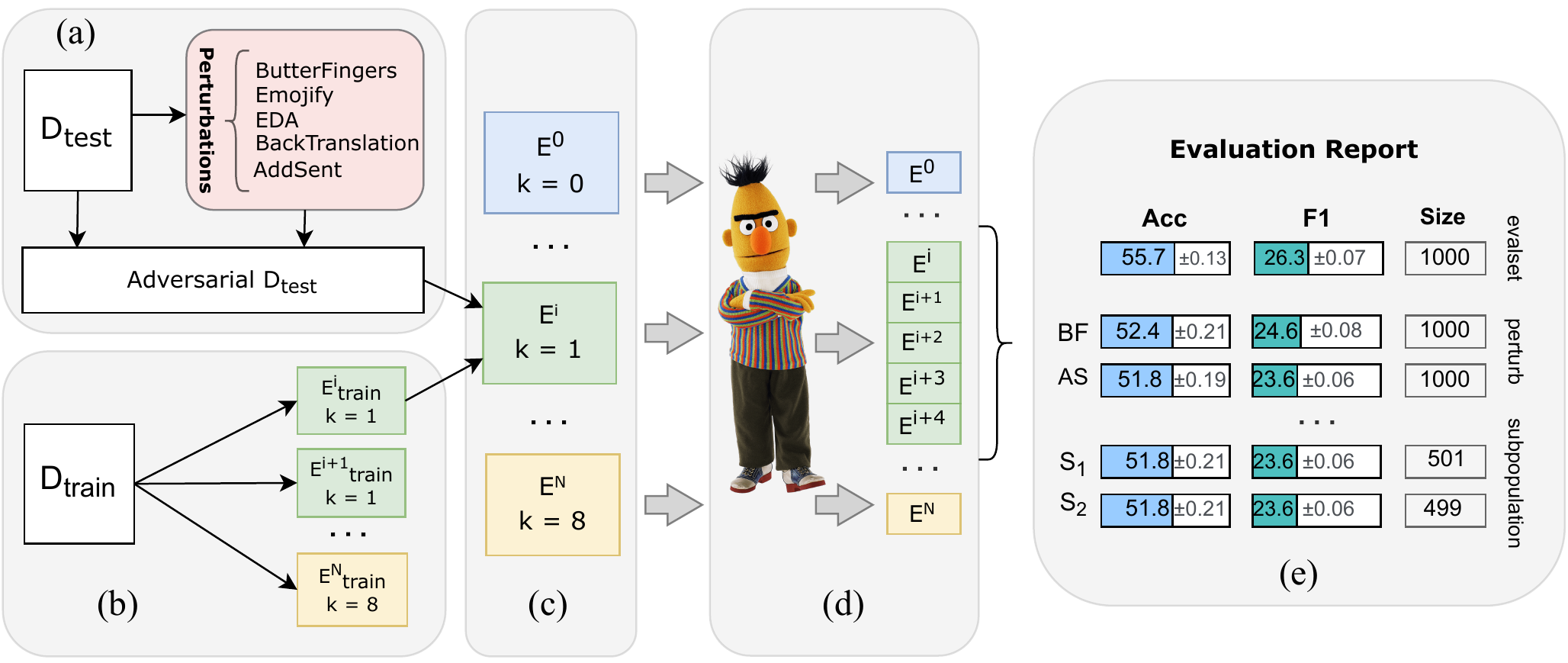}
\caption{Overview of the TAPE's design. \textbf{(a)} $\mathcal{D}_{test}$ is passed to the adversarial framework (\S~\ref{sec:transformation_framework}) to create the adversarial test $\mathcal{D}^{A}_{test}$ that includes the original and adversarial examples. \textbf{(b)} We randomly sample $5$ sets of demonstration examples from $\mathcal{D}_{train}$ for each $k \in \{1, 4, 8\}$. In the zero-shot scenario, we skip this stage. \textbf{(c)} After that, we merge the demonstrations, when applicable, with the examples from $\mathcal{D}^{A}_{test}$ to construct evaluation episodes $\mathcal{E}^{N}_k$. \textbf{(d)} Each $\mathcal{E}^{N}_k$ is used to obtain predictions from the model. \textbf{(e)} The performance is summarized in a diagnostic evaluation report. BF -- \textsc{ButterFingers}, AS -- \textsc{AddSent}, S -- subpopulation. }
\label{fig:eval_setup}
\end{figure*}

\noindent\textbf{CheGeKa.}
\label{chegeka}
The CheGeKa game\footnote{\href{https://en.wikipedia.org/wiki/What\%3F_Where\%3F_When\%3F}{en.wikipedia.org/wiki/what\_where\_when}} setup is similar to Jeopardy, where the player should answer questions based on wit and common sense knowledge. We directly contacted the authors of Russian Jeopardy!~\cite{https://doi.org/10.48550/arxiv.2112.02325} and asked about including their training and private test sets in our benchmark. The task is to provide a free response given a question and the question category.

\begin{itemize}[noitemsep,leftmargin=1.em]
\item \textbf{Question:} \textit{``Imenno on napisal muzyku k opere Turandot.''} (It was he who composed the music for the opera "Turandot".)
\item \textbf{Category:} \textit{``Komediya del' arte''} (Commedia dell'arte)
\item \textbf{Answer:} \textit{``Puchchini''} (Puccini)
\end{itemize}

\subsection{Ethical Judgments}
\label{ethics}
There is a multitude of approaches to evaluating ethics in machine learning. The \textbf{Ethics} dataset for Russian is created from scratch for the first time, relying on the design compatible with~\citet{hendrycks2021aligning}. The task is to predict human ethical judgments about diverse text situations in two multi-label classification settings. The first one is to identify the presence of concepts in normative ethics, such as \emph{virtue, law, moral, justice, and utilitarianism} (\textbf{Ethics$_1$}). The second one is to evaluate the positive or negative implementation of these concepts with binary categories (\textbf{Ethics$_2$}). 

The composition of the dataset is conducted in a semi-automatic mode. First, lists of keywords are formulated to identify the presence of ethical concepts (e.g., \textit{``kill'', ``give'', ``create''}, etc.). The collection of keywords includes the automatic collection of synonyms using the semantic similarity tools of the RusVectores project~\cite{KutuzovKuzmenko2017}. After that, the news and fiction sub-corpora of the Taiga corpus~\cite{shavrina2017methodology} are filtered to extract short texts containing these keywords. Each text is annotated via Toloka as documented in Appendix~\ref{app:data_collection}.

\begin{itemize}[noitemsep,leftmargin=1.em]
\item \textbf{Text:} \textit{``Pechen'kami sobstvennogo prigotovleniya nagradila 100-letnyaya Greta Plokh malysha, kotoryy pomog yey pereyti cherez ozhivlennoye shosse po peshekhodnomu perekhodu.''} (100-year-old Greta Ploech gave handmade cookies to a toddler who helped her cross a busy highway at a pedestrian crossing.)
\item \textbf{Labels$_1$:} \cmark (Virtue) \xmark (Law) \xmark (Moral) \cmark (Justice) \cmark (Utilitarianism)
\item \textbf{Labels$_2$:} \cmark (Virtue) \cmark (Law) \cmark (Moral) \cmark (Justice) \cmark (Utilitarianism)%
\end{itemize}

\section{Design}

\subsection{Evaluation Principles}
\label{sec:evaluation_principles}
This section outlines our evaluation principles that are based on methodological recommendations and open research questions discussed by~\citet{bragg2021flex,bowman-dahl-2021-will,beltagy-etal-2022-zero}, including sample size design, varying number of shots, reporting variability, diagnostic performance analysis, and adversarial test sets. \autoref{fig:eval_setup} describes the TAPE's design.

\vspace{0.1em}
\noindent\textbf{Data Sampling.} Each task in our benchmark consists of a training set $\mathcal{D}_{train}$ with labeled examples and a test set $\mathcal{D}_{test}$. The splits are randomly sampled, except for \textbf{RuOpenBookQA}, \textbf{RuWorldTree}, and \textbf{CheGeKa}, where we use the original splits. We keep the dataset size up to $1$k and purposefully include imbalanced data for the text classification tasks.

\vspace{0.1em}
\noindent\textbf{No extra data.} We do not provide validation sets nor any additional unlabeled data to test the zero-shot and few-shot generalization capabilities of LMs~\cite{bao2019few,tam-etal-2021-improving}. 
 
\vspace{0.1em}
\noindent\textbf{Number of shots.} We consider $k \in \{1, 4, 8\}$ for few-shot evaluation to account for sensitivity to the number of shots. We also include zero-shot evaluation, which can be a strong baseline and simulate scenarios where no supervision is available.

\vspace{0.1em}
\noindent\textbf{Episode sampling.} We provide $5$ episodes in each $k$-shot setting $k \in \{1, 4, 8\}$ and report standard deviation over the episodes to estimate the variability due to the selection of demonstrations~\cite{schick-schutze-2021-just}. Each episode $\mathcal{E}^i = (\mathcal{E}^i_{train_k}, \mathcal{D}^A_{test})$ consists of $k$ demonstration examples $\mathcal{E}^i_{train_k}$ randomly sampled from $\mathcal{D}_{train}$ with replacement, and a single test $\mathcal{D}^A_{test}$ acquired via the combination of original and adversarial test data.

\paragraph{Subpopulations.} Subpopulations~\cite{goel-etal-2021-robustness} are utilized for fine-grained performance analysis w.r.t. such properties of $\mathcal{D}_{test}$ as length, domain, and others.

\vspace{0.1em}
\noindent\textbf{Robustness.} LMs are susceptible to adversarial examples, purposefully designed to force them output a wrong prediction given a modified input \cite{ebrahimi-etal-2018-hotflip, ijcai2018-585, jia-liang-2017-adversarial}. We analyze the LMs' robustness to different types of adversarial data transformations. Here, each $\mathcal{E}^i_{train_k}$ corresponds to $T+1$ test variations, including the original $\mathcal{D}_{test}$ and $T$ adversarial test sets $\mathcal{D}^{A}_{test}$, acquired through the modification of $\mathcal{D}_{test}$. $T$ depends on the dataset and can be adjusted based on the user's needs.

\subsection{Adversarial Framework} \label{sec:transformation_framework}
\subsubsection{Attacks and Perturbations}
\label{subsec:transformations}
\begin{table*}[t!]
    \centering
    \resizebox{\textwidth}{!}{%
    \begin{tabular}{cllp{0.6\linewidth}cc}
        \toprule
          & \bf Type & \bf Name & \bf Example &  \bf Adv. threshold & \bf BERTScore  \\
         \midrule
         \multirow{2}{*}{\resizebox{0.6em}{!}{\rotatebox{90}{Word-l.}}} & Spelling & \textsc{ButterFingers} & \textcolor{cb-green-sea}{\large $\bullet$} \hspace{0em} This is a se\colorbox{cb-grey}{m}tence \colorbox{cb-grey}{r}o test t\colorbox{cb-grey}{j}e code & 0.15 & 82.72 \\\cmidrule{3-6}
         & Modality & \textsc{Emojify} & \textcolor{cb-green-sea}{\large $\bullet$} \hspace{0em} This is a sentence to test the \colorbox{cb-grey}{\codeemoji}\ & 0.4 & 84.27 \\
         \midrule
         \multirow{5}{*}{\resizebox{0.6em}{!}{\rotatebox{90}{Sentence-level}}} 
         & \multirow{2}{*}{Random} & \textsc{EDA$_\text{delete}$} & \textcolor{cb-green-sea}{\large $\bullet$} \hspace{0em} This \colorbox{cb-grey}{\textcolor{cb-grey}{i}} a sentence to test the code & 0.3 & 96.16 \\
         & & \textsc{EDA$_\text{swap}$} & \textcolor{cb-green-sea}{\large $\bullet$} \hspace{0em} \colorbox{cb-grey}{code} is a sentence to test \colorbox{cb-grey}{This} & 0.3 & 93.99 \\\cmidrule{3-6}
         & Paraphrasis &  \textsc{BackTranslation} & \textcolor{cb-green-sea}{\large $\bullet$}  \hspace{0em}  \colorbox{cb-grey}{This sentence tests the code} & 0.5 & 95.38 \\\cmidrule{3-6}
          & Distraction & \textsc{AddSent} &  \textcolor{cb-salmon-pink}{\large $\bullet$}  \hspace{0em} This is a sentence to test the code \colorbox{cb-grey}{, if you want to delete it} & 0.5 & 92.95 \\ 
         \bottomrule
    \end{tabular}}
    \caption{Examples of the TAPE's adversarial attacks and perturbations. The examples are given for the English sentence ``\textit{This is a sentence used to test the code}'' for illustration purposes. The similarity scores for each transformed sentence are given in percent. \textcolor{cb-green-sea}{\large $\bullet$} -- perturbations, \textcolor{cb-salmon-pink}{\large $\bullet$} -- adversarial attacks.} 
    \label{tab:transformations}
\end{table*}

\autoref{tab:transformations} summarizes the TAPE's adversarial attacks and perturbations based on the generally accepted typology~\cite{zhang2020adversarial,wang2021measure}.

\paragraph{Word-level Perturbations.} Word-level perturbations utilize several strategies to perturb tokens, ranging from imitation of typos~\cite{jin2020bert} to synonym replacement~\cite{wei-zou-2019-eda}. We consider the following:

\noindent \underline{Spelling}. \textsc{ButterFingers} is the typo-based perturbation that adds noise to data by mimicking spelling mistakes made by humans through character swaps based on their keyboard distance.

\noindent\underline{Modality}. \textsc{Emojify} replaces the input words with the corresponding emojis, preserving their original meaning. A few authors have manually validated translations of the English emoji dictionary.

\paragraph{Sentence-level Perturbations.} In contrast to word-level perturbations, sentence-level perturbation techniques affect the syntactic structure:

\noindent\underline{Random}. Easy Data Augmentation (\textsc{EDA};~\citealp{wei-zou-2019-eda}) have proved to be efficient in fooling LMs on text classification tasks. We use two \textsc{EDA} configurations: swapping words (\textsc{EDA$_\text{swap}$}) and deleting tokens (\textsc{EDA$_\text{delete}$}).
    
\noindent\underline{Paraphrasis}. \textsc{BackTranslation}~\cite{yaseen-and-langer-backtranslation-ner} allows to generate linguistic variations of the input without changing named entities. We use the OpusMT model\footnote{\href{https://huggingface.co/Helsinki-NLP}{hf.co/Helsinki-NLP/opus-mt}} to translate the input text into English and back to Russian.

\noindent\underline{Distraction}. \textsc{AddSent} is an adversarial attack that generates extra words or sentences with the help of a generative text model. We pass the input to the mGPT\footnote{\href{https://huggingface.co/THUMT/mGPT}{hf.co/THUMT/mGPT}} LM and generate continuations with the sampling strategy. In the multiple-choice QA tasks, we replace one or more incorrect answers with their generated alternatives.

\subsubsection{Data Curation}
Adversarial perturbations and attacks are efficiently utilized to exploit weaknesses in LMs~\cite{goel-etal-2021-robustness}. At the same time, popular techniques may distort semantic meanings or generate invalid adversarial examples~\cite{wang2021adversarial}. We aim to address this problem by using: \emph{(i)} adversarial probability thresholds, \emph{(ii)} task-specific constraints, and \emph{(iii)} semantic filtering.

\paragraph{Probability thresholds.} The degree of the input modification can be controlled with an adversarial probability threshold, which serves as the hyperparameter. The higher the probability, the more the input gets modified. The thresholds used in our experiments are specified in~\autoref{tab:transformations}.

\paragraph{Constraints.} The TAPE's attacks and perturbations do not drastically change the input's meaning. Despite this, we consider using rule-based constraints that keep the linguistic structure and task-specific aspects unchanged (see~\autoref{tab:constraints} in~\autoref{app:constraints}). For instance, it is crucial to leave named entities in the QA tasks untouched or not modify the syntactic structure and anaphors when perturbing the \textbf{Winograd} examples.

\paragraph{Semantic filtering.} We follow~\citeauthor{wang2021adversarial} on filtering the adversarial examples with BERTScore\footnote{\href{https://huggingface.co/bert-base-multilingual-cased}{hf.co/bert-base-multilingual-cased}}~\cite{zhang2019bertscore}, a BERT-based text similarity metric~\cite{devlin-etal-2019-bert}. We measure the semantic similarity between the original input and adversarial output and keep examples with the highest similarity score. In cases when the score is lower than a specified threshold, we iteratively decrease the adversarial probability threshold and re-score the new adversarial examples.

\begin{table*}[t]
    \centering
    \resizebox{\textwidth}{!}{%
    \begin{tabular}{llcccccp{0.27\linewidth}cclp{0.1\linewidth}}
        \toprule
         & \bf Dataset & \bf BF & \bf EMJ & \bf EDA & \bf BT & \bf AS & \bf Subpopulations & \bf $|\mathcal{D}_{train}|$ & \bf $|\mathcal{D}_{test}|$ & \bf Metrics & \bf Domain \\
        \midrule
        \multirow{7}{*}{\rotatebox{90}{Classification}}
           & \multirow{4}{*}{Winograd} 
                & \multirow{4}{*}{\cmark}  & \multirow{4}{*}{\cmark} & \multirow{4}{*}{\cmark}& \multirow{4}{*}{\xmark} & \multirow{4}{*}{\cmark} & \vspace{-0.7em} \begin{itemize}[noitemsep,topsep=0pt,leftmargin=1em]
                    \item[\textcolor{cb-rose}{\large $\bullet$}] Gender, Number
                    \item[\textcolor{cb-purple}{\large $\bullet$}]  NumCandidates
                    \item[\textcolor{cb-burgundy}{\large $\bullet$}]  Class distribution
                    \item[\textcolor{cb-blue}{\large $\bullet$}] Length
                \end{itemize}\baselineskip=0em & 60 & 986 & F1 / Acc. &  Misc. \\\cmidrule{3-12}
            
           & \multirow{3}{*}{Ethics$_{1/2}$} 
                & \multirow{3}{*}{\cmark}  &  \multirow{3}{*}{\cmark} & \multirow{3}{*}{\cmark} & \multirow{3}{*}{\cmark} & \multirow{3}{*}{\cmark} &\vspace{-0.7em}
                \begin{itemize}[noitemsep,topsep=0pt,leftmargin=1em]
                    \item[\textcolor{cb-blue-green}{\large $\bullet$}] Domain 
                    \item[\textcolor{cb-blue}{\large $\bullet$}] Length
                    \item[\textcolor{cb-green-sea}{\large $\bullet$}] Diversity, Readability
                \end{itemize}\baselineskip=0em & 59/58 & 1k/1k & F1 / Acc. & News, fiction books \\
        \midrule
        \multirow{13}{*}{\rotatebox{90}{Question Answering}}
            & \multirow{3}{*}{RuWorldTree} 
                &  \multirow{3}{*}{\cmark} & \multirow{3}{*}{\cmark} & \multirow{3}{*}{\cmark} & \multirow{3}{*}{\cmark} & \multirow{3}{*}{\cmark} &
                \vspace{-0.7em}
                \begin{itemize}[noitemsep,topsep=0pt,leftmargin=1em]
                    \item[\textcolor{cb-salmon-pink}{\large $\bullet$}] Answer Category
                    \item[\textcolor{cb-blue-green}{\large $\bullet$}] ExamName,SchoolGrade
                    \item[\textcolor{cb-blue}{\large $\bullet$}] Length
                    \item[\textcolor{cb-green-sea}{\large $\bullet$}] Diversity, Readability
                \end{itemize}\baselineskip=0em & 47 & 629 & F1 / Acc. & Elementary science exams \\\cmidrule{3-12}
    
            & \multirow{3}{*}{RuOpenBookQA} 
                 & \multirow{3}{*}{\cmark}  & \multirow{3}{*}{\cmark} & \multirow{3}{*}{\cmark} & \multirow{3}{*}{\cmark} & \multirow{3}{*}{\cmark} & \vspace{-0.7em}
                \begin{itemize}[noitemsep,topsep=0pt,leftmargin=1em]
                    \item[\textcolor{cb-salmon-pink}{\large $\bullet$}] Answer Category
                    \item[\textcolor{cb-blue}{\large $\bullet$}] Length
                    \item[\textcolor{cb-green-sea}{\large $\bullet$}] Diversity, Readability
                \end{itemize}\baselineskip=0em & 48 & 500 & F1 / Acc. & Elementary science exams \\\cmidrule{3-12}
            
            & \multirow{3}{*}{MultiQ} 
                & \multirow{3}{*}{\cmark} & \multirow{3}{*}{\cmark} & \multirow{3}{*}{\cmark} & \multirow{3}{*}{\cmark} & \multirow{3}{*}{\cmark} &\vspace{-0.7em}\begin{itemize}[noitemsep,topsep=0pt,leftmargin=1em]
                    \item[\textcolor{cb-salmon-pink}{\large $\bullet$}] Answer Category
                    \item[\textcolor{cb-blue}{\large $\bullet$}] Length
                    \item[\textcolor{cb-green-sea}{\large $\bullet$}] Diversity, Readability
                \end{itemize}\baselineskip=0em & 64 & 1k & F1 / EM & Wikipedia \\\cmidrule{3-12}
            
            & \multirow{3}{*}{CheGeKa} 
                & \multirow{3}{*}{\cmark} & \multirow{3}{*}{\cmark} & \multirow{3}{*}{\cmark} & \multirow{3}{*}{\cmark} & \multirow{3}{*}{\cmark} &\vspace{-0.7em}\begin{itemize}[noitemsep,topsep=0pt,leftmargin=1em]
                    \item[\textcolor{cb-salmon-pink}{\large $\bullet$}] Answer Category
                    \item[\textcolor{cb-blue}{\large $\bullet$}] Length
                    \item[\textcolor{cb-green-sea}{\large $\bullet$}] Diversity, Readability
                \end{itemize}\baselineskip=0em & 49 & 520 & F1 / EM & General domain \\
        \bottomrule
    \end{tabular}%
    }
    \caption{Summary of the TAPE benchmark. Transformations: BF -- \textsc{ButterFingers}, EMJ -- \textsc{Emojify}, AS --  \textsc{AddSent}, \textsc{EDA} includes  \textsc{EDA$_\text{swap}$} and \textsc{EDA$_\text{delete}$}. Subpopulations: \textcolor{cb-rose}{\large $\bullet$} -- Morphology, \textcolor{cb-burgundy}{\large $\bullet$} -- Class Distribution, \textcolor{cb-blue-green}{\large $\bullet$} -- Domain, \textcolor{cb-salmon-pink}{\large $\bullet$} -- Answer Category, \textcolor{cb-blue}{\large $\bullet$} -- Length, \textcolor{cb-purple}{\large $\bullet$} -- Number of Candidates, \textcolor{cb-green-sea}{\large $\bullet$} -- Text Statistics.} 
    \label{tab:stats}
\end{table*}

\section{Baselines}
\subsection{Non-neural Baselines}
We use two models from the scikit-learn library~\cite{pedregosa2011scikit} as non-neural baselines for classification (\textbf{Ethics$_{1/2}$} and \textbf{Winograd}) and multiple-choice QA tasks (\textbf{RuOpenBookQA} and \textbf{RuWorldTree}). The baselines are fit on the corresponding $\mathcal{D}_{train}$ and evaluated on $\mathcal{D}_{test}$.

\vspace{0.5mm}
\noindent \textbf{Random} is a simple data-agnostic baseline that samples predictions uniformly from the set of target classes in a given task.

\noindent \textbf{Linear} is a logistic regression classifier over TF-IDF~\cite{Salton1973OnTS} $N$-grams within the range $N \in [1; 4]$. The classifier is trained on top-$150$k features with default L2-regularization hyperparameters.

\subsection{Neural Baselines}
We run zero-shot and few-shot evaluation of Russian GPT3\footnote{\href{https://github.com/ai-forever/ru-gpts}{github.com/ai-forever/ru-gpts}} LMs available under the HuggingFace library~\cite{wolf-etal-2020-transformers}. We consider three model versions: ruGPT3$_\text{S}$\footnote{\href{https://huggingface.co/sberbank-ai/rugpt3small_based_on_gpt2}{hf.co/sberbank-ai/rugpt3small}} (125M), ruGPT3$_\text{M}$\footnote{\href{https://huggingface.co/sberbank-ai/rugpt3medium_based_on_gpt2}{hf.co/sberbank-ai/rugpt3medium}} (350M), and ruGPT3$_\text{L}$\footnote{\href{https://huggingface.co/sberbank-ai/rugpt3large_based_on_gpt2}{hf.co/sberbank-ai/rugpt3large}} (760M).

\paragraph{Perplexity-based evaluation.} We consider the setting where the classification and multiple-choice tasks are formulated in natural language as a cloze-style prompt template: \textbf{Winograd}, \textbf{Ethics$_{1/2}$}, \textbf{RuWorldTree} and \textbf{RuOpenBookQA}. We provide examples of the prompt templates for each task in~\autoref{app:prompt_formats}. After filling in each possible target class or choice, we compute the per-token cross-entropy loss, which is reduced to negative log-probability due to one-hot encoding of the target tokens. The most probable string has the lowest sum of negative log probabilities of its tokens normalized over the total number of tokens in the input, as specified in~\autoref{eqn:ppl}.

\vspace{-0.5cm}
\begin{equation}
  \text{PPL}(t) = \exp(-\frac{1}{|t|}\sum_{i=0}^{|t|} 
log_{p_{\theta}}(x_i|x_{<i})),
\label{eqn:ppl}
\end{equation}

\noindent where $t$ is the input prompt and $|t|$ is the length of the prompt in tokens. The choice relies on our preliminary experiments, where instead, the most probable string is chosen based on the lowest sum of negative log probabilities of the prompt's tokens. However, the latter approach has shown worse results on the subsets of the training sets.

\paragraph{Zero-shot and few-shot generation.} Text generative baselines are of the greatest interest for tasks that can not be solved by the perplexity-based approach: \textbf{CheGeKA} and \textbf{MultiQ}. Here, we generate the answer given the corresponding task prompt (see~\autoref{app:prompt_formats}) with nucleus sampling ($top_p=0.8$). The choice of the strategy and hyperparameters is based on the grid search\footnote{Beam search: number of beams $\{1,2,5,10,50\}$; Nucleus sampling: $top_p \in \{0.5,0.8,0.9,0.95-0.99\}$. We also experimented with top-k sampling and greedy decoding.} on a subset of the corresponding $\mathcal{D}_{train}$. The output is limited by $100$/$200$ (\textbf{CheGeKA}) and $400$/$800$ (\textbf{MultiQ}) tokens in zero-shot/few-shot settings, respectively.

\subsection{Human Baselines}
The human evaluation is run via Toloka. Access to the annotation projects is granted to annotators certified as Russian native speakers. Each project consists of an unpaid training stage, control examples for monitoring annotation quality\footnote{Control examples are commonly used on Toloka for filtering out results from bots, cheaters, and low-performing annotators. The examples are manually selected by a few authors from the $\mathcal{D}_{train}$ and guaranteed to be unambiguous.}, and the main annotation stage. The annotator is given detailed instruction with a task description, annotation examples, and corresponding explanations. The instruction is linked to training and main annotation stages and available any time. Annotators who get less than $60$\% of the training examples correct on average do not qualify for the main stage. Each qualified annotator receives a page with a certain number of examples for annotation, one of which is a control one. The inter-annotator agreement (IAA) is based on the Dawid-Skene aggregation model\footnote{\href{https://toloka.ai/docs/guide/concepts/result-aggregation.html}{toloka.ai/docs/result-aggregation}}~\cite{dawid1979maximum}, available directly from Toloka. Details on the annotation process, inter-annotator agreement rates, hourly pay rate, average response time, the overall project cost, annotation instructions, and examples of web interface are fully documented in~\autoref{sec:appendix_annotation}.

\begin{table*}[!ht]
    \centering
    \resizebox{\textwidth}{!}{%
    \begin{tabular}{lccccccccc}
    \toprule
    \bf Model & \bf \textit{k}-shot & \bf Winograd & \bf RuWorldTree & \bf RuOpenBookQA & \bf MultiQ & \bf CheGeKa & \bf Ethics$_1$ & \bf Ethics$_2$ & \bf Avg \\
    \midrule
    \multirow{4}{*}{ruGPT3$_\text{S}$} 
        & 0 & 38.6 / 57.9 & 34.1 / 34.0 & 33.9 / 34.0 & 0.17 / 0.0 & \underline{1.0} / 0.0 & 34.1 / 55.5 & 48.5 / 60.9 & 27.24 / 34.64 \\
        & 1 & 36.7 {\tiny$\pm$0.0} / 58.0 {\tiny$\pm$0.0} & 33.5 {\tiny$\pm$0.4} / 33.5 {\tiny$\pm$0.4} & 34.9 {\tiny$\pm$0.8} / 35.0 {\tiny$\pm$0.8} & 0.13 {\tiny$\pm$0.04} / 0.0 {\tiny$\pm$0.0} & 0.52 {\tiny$\pm$0.14} / 0.0 {\tiny$\pm$0.0} & 14.3 {\tiny$\pm$11.1} / 59.9 {\tiny$\pm$5.8} & 49.6 {\tiny$\pm$7.1} / 60.8 {\tiny$\pm$5.7} & 24.24 / 35.33\\
        & 4 & 41.0 {\tiny$\pm$5.3} / 54.8 {\tiny$\pm$3.9} & 34.3 {\tiny$\pm$0.9} / 34.3 {\tiny$\pm$0.9} & 34.6 {\tiny$\pm$0.6} / 34.6 {\tiny$\pm$0.6} & 0.09 {\tiny$\pm$0.06} / 0.0 {\tiny$\pm$0.0} & 0.62 {\tiny$\pm$0.34} / 0.65 {\tiny$\pm$0.29}  & 23.5 {\tiny$\pm$13.8} / 54.3 {\tiny$\pm$12.3} & 56.7 {\tiny$\pm$10.8} / 61.3 {\tiny$\pm$5.0} & 27.44 / 34.31\\
        & 8 & 36.9 {\tiny$\pm$0.4} / 54.8 {\tiny$\pm$6.5} & 35.1 {\tiny$\pm$0.9} / 35.1 {\tiny$\pm$0.8} & 33.6 {\tiny$\pm$0.5} / 33.7 {\tiny$\pm$0.5} & 0.12 {\tiny$\pm$0.22} / 0.0 {\tiny$\pm$0.0} & 0.35 {\tiny$\pm$0.26} / 0.35 {\tiny$\pm$0.26} & 32.4 {\tiny$\pm$6.9} / 50.4 {\tiny$\pm$4.0} & \underline{63.7} {\tiny$\pm$9.5} / 61.7 {\tiny$\pm$4.2} & \underline{29.04} / 33.56 \\ 
    \midrule
    \multirow{4}{*}{ruGPT3$_\text{M}$} 
        & 0 & 39.5 / 57.2 & 38.0 / 38.0 & 34.8 / 34.8 & \underline{0.18} / 0.0 & 0.75 / 0.0 & 2.4 / 68.3 & 5.3 / 44.1 & 17.29 / 34.71 \\
        & 1 & 36.7 {\tiny$\pm$0.0} / 58.0 {\tiny$\pm$0.0} & 36.3 {\tiny$\pm$0.6} / 36.3 {\tiny$\pm$0.6} & 36.7 {\tiny$\pm$0.8} / 36.8 {\tiny$\pm$0.8} & 0.11 {\tiny$\pm$0.03} / 0.0 {\tiny$\pm$0.0} & 0.47 {\tiny$\pm$0.11} / 0.0 {\tiny$\pm$0.0} & 14.3 {\tiny$\pm$11.1} / 59.9 {\tiny$\pm$5.8} & 49.6 {\tiny$\pm$7.1} / 60.8 {\tiny$\pm$5.7} & 24.9 / 35.99\\
        & 4 & 39.7 {\tiny$\pm$4.1} / 57.9 {\tiny$\pm$0.6} & 38.9 {\tiny$\pm$1.0} / 38.8 {\tiny$\pm$1.0} & 36.3 {\tiny$\pm$1.3} / 36.4 {\tiny$\pm$1.3} & 0.12 {\tiny$\pm$0.18} / 0.0 {\tiny$\pm$0.0} & 0.85 {\tiny$\pm$0.52} / \underline{0.92} {\tiny$\pm$0.53} & 14.9 {\tiny$\pm$16.0} / 58.6 {\tiny$\pm$13.1} & 45.5 {\tiny$\pm$13.7} / 59.0 {\tiny$\pm$7.3} & 25.57 / 35.74\\
        & 8 & 39.0 {\tiny$\pm$5.9} / 54.3 {\tiny$\pm$6.5} & 40.1 {\tiny$\pm$0.5} / 40.1 {\tiny$\pm$0.5} & 35.7 {\tiny$\pm$0.9} / 35.8 {\tiny$\pm$1.0} & 0.1 {\tiny$\pm$0.15} / 0.0 {\tiny$\pm$0.0} & 0.42 {\tiny$\pm$0.19} / 0.42 {\tiny$\pm$0.19} & 19.6 {\tiny$\pm$10.6} / 60.1 {\tiny$\pm$5.1} & 57.7 {\tiny$\pm$9.0} / 64.2 {\tiny$\pm$2.1} & 27.78 / 36.83 \\
    \midrule
    \multirow{4}{*}{ruGPT3$_\text{L}$} 
        & 0 & 39.2 / 55.5 & 40.8 / 40.7 & 39.6 / 39.6 & 0.14 / 0.0 & 0.57 / 0.0 & 0.3 / \underline{68.6} & 8.0 / 44.9 & 18.15 / 35.73 \\
        & 1 & 36.7 {\tiny$\pm$0.0} / 58.0 {\tiny$\pm$0.0} & 38.2 {\tiny$\pm$0.8} / 38.2 {\tiny$\pm$0.8} & \underline{39.8} {\tiny$\pm$1.0} / \underline{39.8} {\tiny$\pm$1.1} & 0.12 {\tiny$\pm$0.4} / 0.0 {\tiny$\pm$0.0} & 0.47 {\tiny$\pm$0.11} / 0.0 {\tiny$\pm$0.0} & 14.3 {\tiny$\pm$11.1} / 59.9 {\tiny$\pm$5.8} & 49.6 {\tiny$\pm$7.1} / 60.8 {\tiny$\pm$5.7} & 25.6 / 36.69\\
        & 4 & 38.8 {\tiny$\pm$2.9} / \underline{58.4} {\tiny$\pm$0.5} & 40.9 {\tiny$\pm$0.8} / 40.8 {\tiny$\pm$0.8} & 39.6 {\tiny$\pm$0.8} / 39.6 {\tiny$\pm$0.8} & 0.1 {\tiny$\pm$0.14} / 0.0 {\tiny$\pm$0.0} & 0.54 {\tiny$\pm$0.44} / 0.54 {\tiny$\pm$0.4} & 11.8 {\tiny$\pm$12.4} / 60.4 {\tiny$\pm$11.1} & 44.5 {\tiny$\pm$15/9} / 59.6 {\tiny$\pm$8.1} & 25.1 / 36.96\\
        & 8 & 39.2 {\tiny$\pm$3.2} / 57.6 {\tiny$\pm$1.5} & \underline{41.9} {\tiny$\pm$0.7} / \underline{41.9} {\tiny$\pm$0.7} & 38.8 {\tiny$\pm$1.5} / 38.9 {\tiny$\pm$1.5} & 0.0 {\tiny$\pm$0.0} / 0.0 {\tiny$\pm$0.0} & 0.54 {\tiny$\pm$0.4} / 0.54 {\tiny$\pm$0.4} & 23.3 {\tiny$\pm$8.3} / 58.1 {\tiny$\pm$3.4} & 52.0 {\tiny$\pm$8.5} / 63.7 {\tiny$\pm$2.4} & 28.33 / \underline{37.12} \\
    \midrule
    Random 
        & \xmark & \underline{51.7} / 51.3 & 26.0 / 26.1 & 21.4 / 21.6 & \xmark & \xmark & \underline{36.5} / 47.9 & 54.1 / 52.0 & \xmark \\ %
    Linear
        & \xmark & 36.7 / 58.0 & 14.3 / 25.8 & 11.4 / 25.2 & \xmark & \xmark & 0.0 / \underline{68.6}  & 49.0 / \underline{65.3} & \xmark \\ %
    Human 
        & \xmark & \bf 86.6 / 87.0 & \bf 83.8 / 83.7 & \bf 87.5 / 86.5 & \bf 92.8 / 91.0 &   \bf 71.9 / 64.5 & \bf 51.0 / 52.9 & \bf 73.0 / 67.6 & \bf 78.09 / 76.17 \\
    \bottomrule
    \end{tabular}}
    \caption{Performance results of the non-neural, neural, and human baselines on the original test sets. Metrics: F1-score/accuracy (EM). The best score is put in bold, the second best is underlined.}
    \label{tab:eval_sets_results}
\end{table*}
\subsection{Metrics} 
We evaluate the baseline performance with macro-averaged F1 and accuracy scores for the classification (\textbf{Winograd}, \textbf{Ethics$_{1/2}$}) and multiple-choice QA tasks (\textbf{RuOpenBookQA}, \textbf{RuWorldTree}). F1-score and exact match (EM) are used for the open-domain (\textbf{CheGeKa}) and multi-hop QA tasks (\textbf{MultiQ}) -- obtaining such metrics for generative and sequence-to-sequence tasks provides a comparable yet strict setup. The effectiveness of the perturbations and adversarial attacks is measured with the attack success rate (ASR;~\citealp{wang2021adversarial}) computed as the percentage of the correct predictions that are changed after the perturbation or attack is applied.

\section{Results}

\subsection{Generalization Evaluation} \label{sec:generalization}
\autoref{tab:eval_sets_results} presents the zero-shot and few-shot performance results of the non-neural, neural, and human baselines on the original $\mathcal{D}_{test}$ sets.

\vspace{0.1em} \noindent\textbf{Classification and multiple-choice tasks.} The zero-shot evaluation provides a strong baseline, matching or exceeding the few-shot performance on \textbf{Winograd} ($k \in \{1,4\}$) and \textbf{Ethics$_1$} ($k \in \{1,4,8\}$). The zero-shot performance is similar among the models despite their size (\textbf{Winograd}), or it can steadily improve (\textbf{RuWorldTree}, \textbf{RuOpenBookQA}) and significantly drop when the model size increases (\textbf{Ethics$_{1/2}$}). We observe that introducing more examples increases variability on the imbalanced classification tasks (\textbf{Winograd}, \textbf{Ethics$_{1/2}$}) and leads to performance degradation, specifically for ruGPT$_\text{S}$. Furthermore, the performance degenerates into constant predictions, which is indicated by the significant difference in accuracy and F1 scores on the (\textbf{Winograd}) and (\textbf{Ethics$_{1/2}$}) tasks. In particular, the LMs predict the negative label for about 97\% of the \textbf{Winograd} samples in the zero-shot setting. In the few-shot, however, the number of constant predictions is reduced to 80\% ($k \in \{4,8\}$). This result indicates that the demonstrations may help generalize to the task, but the predictions are still affected by the imbalanced classification setting. We also observe that \textbf{Ethics$_{1/2}$} is the most challenging task for both human and neural baselines. The results are sensitive to prompt changing, and human annotators may receive low inter-annotator agreement on the examples due to subjectivity.

\vspace{0.1em}\noindent\textbf{Zero-shot and few-shot generation results.} 
However, approaching the  \textbf{CheGeKa} and \textbf{MultiQ} tasks with zero-shot and few-shot generation results in near-zero generalization performance. Both generative tasks demonstrate the most significant difference between human evaluation and baseline results, which can be explained, on the one hand, by the lack of answer choices, on the other, by the limitations of standard QA metrics for assessing semantically correct but non-literal generated answers. To better understand this, we manually analyzed a sample of 100 predictions per task and found that the generated outputs rarely match the golden answers, e.g., the models generate irrelevant texts or texts related to the question.

\vspace{0.1em} \noindent\textbf{Discussion.} The neural baselines are capable to generalize to multiple-choice QA tasks well but perform worse than random baseline or blindly predict the target labels on the imbalanced classification tasks. Our results are consistent with~\citet{DBLP:journals/corr/abs-2112-10668} in that: \emph{(i)} the few-shot evaluation results may rely heavily on the input prompts, \emph{(ii)} it is difficult for GPT-style LMs to perform judgments on social value tasks in the zero-shot and few-shot settings, and \emph{(iii)} the few-shot results on some tasks are worse than zero-shot, meaning that LMs are not able to utilize given demonstrations for solving them. We also observe the \underline{negative result}: zero-shot and few-shot generation baselines may receive near-zero performance on the open-domain and extractive QA tasks.

\begin{table}[t!]
\centering
\resizebox{\columnwidth}{!}{%
\begin{tabular}{lcccccccc}
\toprule
\textbf{Model} &
  \textbf{\textit{k}-shot} &
  \textbf{BF} &
  \textbf{EMJ} &
  \textbf{EDA$_\textsc{swap}$} &
  \textbf{EDA$_\textsc{del}$} &
  \textbf{BT} &
  \textbf{AS} &
  \textbf{Avg} \\ \midrule 
\multirow{4}{*}{ruGPT3$_\textsc{S}$} &
  0 & 53.0  & 37.3  & \underline{33.6}  & 37.9  & 22.3 & 18.5  & 33.77 \\
& 1 & 57.55 & 42.1  & 37.3  & 44.7  & \bf 19.4 & 12.05 & 35.52 \\
& 4 & 58.8  & 35.85 & 39.8  & 47.5  & \underline{19.5} & 8.85  & 35.05 \\
& 8 & 60.45 & 36.65 & 43.45 &  50.25 & 20.0 & 7.2 &  36.33 \\ \midrule

\multirow{4}{*}{ruGPT3$_\textsc{M}$}
& 0 & \bf 50.3  &  38.9  & \bf 33.5  & \bf 41.0 & 20.75 & 14.25 & \bf 33.12 \\
& 1 & 57.4  & 35.25 & 36.05 & 43.55 & 20.7  & 7.55  & 33.41 \\
& 4 & 60.1  & 34.55 & 37.8  & 46.0  & 20.35 & 5.45  & 34.04 \\
& 8 & 61.35 & 34.85 & 38.95 & 48.2  & 21.55 & \underline{3.55}  & 34.74 \\ \midrule 
\multirow{4}{*}{ruGPT3$_\textsc{L}$} 
& 0 & \underline{52.1}  & 36.2  & 35.8  & \underline{43.3}  &  25.25 & 12.65 & 34.21 \\
& 1 & 55.3  & \bf 33.1  & 37.15 & 43.8  & 22.25 & 7.7   & \underline{33.22} \\
& 4 & 59.15 & 34.95 & 39.05 & 46.3  & 21.85 & 7.9   & 34.87 \\
& 8 &  62.0  & \underline{34.4}  & 41.65 & 47.6 & 21.8  & \bf 2.7   & 35.03 \\ \bottomrule
\end{tabular}%
}
\caption{The robustness evaluation results by adversarial perturbation and attack. The ASR values are averaged over the \textbf{RuOpenBookQA} and \textbf{RuWorldTree} tasks. The lower, the better. The best ASR value is put in bold and the second best is underlined. }
\label{tab:asr_by_model}
\end{table}

\subsection{Robustness}
\autoref{tab:asr_by_model} shows the ASR scores for each perturbation and \textit{k}-shot setting averaged over the \textbf{RuWorldTree} and \textbf{RuOpenBookQA} tasks, where the model performance exceeds the random baseline. We observe that the models are more susceptible to simple spelling-based perturbations (\textsc{ButterFingers}), token deletion (\textsc{EDA}$_\textsc{delete}$), and modality changes (\textsc{Emojify}). Paraphrasing the input (\textsc{BackTranslation}) and, especially, addition of the distractors (\textsc{AddSent}), in turn, bear significantly less effect. Most of the perturbations either gain effectiveness as the number of demonstrations increases (\textsc{ButterFingers}, \textsc{EDA}$_\textsc{delete}$, \textsc{EDA}$_\textsc{swap}$) or does not seem to directly depend on the $k$ value (\textsc{EMJ}, \textsc{BackTranslation}). However, the ASR scores for the distraction adversarial attack (\textsc{AddSent}) are inversely proportional to the number of shots. In other words, models are less likely to fall for the generated answer option as the number of demonstrations increases. The model size is not the main factor in model robustness, and the difference between the models is quite subtle. We notice, however, that larger models (ruGPT3$_\textsc{M}$, ruGPT3$_\textsc{L}$) are less affected by the \textsc{AddSent} perturbation and more vulnerable to \textsc{BackTranslation}, in comparison to ru-GPT3$_\textsc{S}$. Nonetheless, the latter proves to be less robust to adversarial perturbations, which is indicated by the higher ASR scores on average, closely followed by the largest model. 

\vspace{0.1em} \noindent\textbf{Discussion.} The neural baselines are most vulnerable to word-level text perturbations: typo-based transformations and modality changes. In terms of the sentence-level perturbations, simple EDA techniques prove to be more effective than paraphrasis and the distraction-based adversarial attack, which can be due to the task-specificity. This is in line with \citet{wang2021adversarial}, who observe typo-based adversarial perturbations as the most effective automatic attack methods.

\subsection{Diagnostic Analysis}
Subpopulation analysis reveals that the GPT-style models exhibit length bias, which is indicated by low performance on longer inputs (see~\autoref{fig:report0}-\autoref{fig:report8} in~\autoref{app:subpopulations}). The effect is observed primarily on the \textbf{RuWorldTree} task, where the models' performance significantly drops on longer texts. As might be expected, for QA tasks, question complexity, determined by school grade or exam name, also affects the model performance. The models tend to better deal with easier questions, which becomes more prominent as the number of demonstrations increases. Seemingly, the readability and lexical diversity of a question, determined by Flesch Reading Ease scores and type-token ratio, respectively, do not affect model performance. Nevertheless, slight increases in performance on more readable and diverse texts are present.

\noindent\textbf{Discussion.} We reveal that the baseline performance depends on the input length. One of the reasons for such behavior can be the limited context window that the models have. \citet{alex2021raft} have previously explored reasoning over long texts in a few-shot setting and their results are consistent with our findings.

\section{Conclusion and Future Work}
Zero-shot and few-shot methods have evolved as a new paradigm in NLP. Addressing the best practices, we introduced TAPE, a text attack and perturbation evaluation benchmark for Russian. TAPE combines the general language understanding evaluation techniques with the greener no-tuning approach, allowing the evaluation of LMs' robustness on complex intellectual tasks. We present six new datasets and a framework for generating adversarial attacks and perturbations, which can also be used as a standalone tool for practical purposes. %

In future, we plan to incorporate more LMs with various architectures and prompting-based methods into the framework. Another direction is to evaluate the cross-lingual generalization capabilities of autoregressive LMs. We hope to encourage the community to foster evaluation of LMs' generalization capacity in non-English languages, leading to the development of more robust and reliable LMs.

\section{Limitations}
\textbf{Performance aggregation.} The well-established GLUE-style benchmarks evaluate systems using mean aggregation over heterogeneous task-specific metrics~\cite{wang-etal-2018-glue,wang2019superglue,wang2021adversarial}. Based on the criticism of this evaluation protocol by the research community~\citep[e.g.,][]{waseem2021disembodied,mishra2021robust,agarwal2021deep}, we recognize that mean aggregation in our case does not account for the nature of the adversarial transformations and attacks and task specifications, such as the task type, domain, and the number of episodes in $\mathcal{D}_{train}$ and $\mathcal{D}_{test}$.

\noindent \textbf{Baseline evaluation.} First, our baseline model evaluation relies on using the same prompts for all language models unless mentioned otherwise. Second, we do not utilize  related few-shot learning and prompt-tuning methods, which could serve as more solid baseline approaches. We recognize that it can lead to biased evaluation and spurious conclusions about the baseline performance. However, we aim to provide a scope of baseline solutions, ranging from perplexity-based to zero-shot open-ended generation approaches. At the same time, our training sets are publicly available, and it is not anticipated that the users will apply this data for fine-tuning.

\noindent \textbf{Human performance.} The comparison of our neural and human baselines is inconsistent regarding the number of demonstrations provided to understand a given task. The zero-shot and few-shot human performance can be comparable to neural LMs' performance when humans would receive $k \in \{0, 1, 4, 8\}$ examples in the annotation training stage~\cite{mukherjee2021few}.

\section{Ethics Statement}
\noindent \textbf{Subjectivity related to ethics.}
Ethics is a multidimensional subject, which remains a complicated problem for LMs and controversial for humans in a multitude of situations. Although our methodology spans general concepts in normative ethics, we acknowledge that it can be challenging to perform \emph{objective} ethical judgments about some situations~\cite{martineau2006types}. For instance, judgments about \emph{law} are based on formal criteria (e.g., the criminal code), \emph{morality} may rely on public sentiment, while \emph{justice} may heavily rely on private sentiment and human worldview. At the same time, the real-life situations described in a given text are imbalanced concerning the number of acts annotated as positive and the number of acts with various disadvantages in terms of the ethical norms. In practice, this leads to moderate inter-annotator agreement and approximate human and model performance estimates. 

\noindent \textbf{Risks related to ethics.} We acknowledge that approaches to evaluating LMs' ability to perform ethical judgments about text situations have been criticized~\cite{talat-etal-2022-machine}. While we use a similar set of ethical concepts~\cite{hendrycks2021aligning}, we collect annotations according to the five criteria that describe the aspects of the annotators' attitude towards the deed. The attitude can be determined by various individual and social aspects. Here, we have analyzed metadata of our Ethics$_1$ annotators available via the Toloka interface. There are $481$ Russian speakers across $16$ different countries, who can be grouped by their age as follows: $18-30$ ($163$ annotators); $30-50$ ($265$ annotators), and $50-78$ ($53$ annotators). Thus, we will further take into account specific risks arising within the annotation process:

\noindent \underline{Social properties:} the diffusion of norms in the Russian-speaking communities has been the object of rapid changes \cite{doi:10.1080/13510347.2021.1928078}. This can be expressed in a shift in attitude towards actions that have different interpretations from the point of view of regional cultural norms, cultures of small peoples, religious norms, and normative behavior for classes of society.

\noindent   \underline{Legal properties:} as the ``legality'' of a deed in a text can change over time, we are sure to see a growing annotation inconsistency in individual examples that reflect societal changes after some years.

\noindent The risks are partially mitigated by the prior training of the annotators and annotator's performance control. Running the annotation experiments from year to year is reasonable to understand possible norm shifts, measuring the variation in annotators' opinions about aspects of the described actions. Furthermore, other data-dependent risks can be indicated, such as genre bias and author's bias in specific publicly available text sources.

\noindent \textbf{Societal impact.} The TAPE's design allows us to alleviate the problems of a large carbon footprint~\cite{bender2021dangers} and keep computational costs accessible to academic and industrial fields~\cite{couldry2020costs}. In particular, our evaluation approach does not consider LMs' fine-tuning and relies on a limited amount of episodes, while the number of attacks and perturbations can be adjusted based on the user's needs. %

\bibliography{custom,anthology}
\bibliographystyle{acl_natbib}

\clearpage
\newpage

\appendix
\setcounter{figure}{0}
\setcounter{table}{0}

\section{General Dataset Statistics}
\label{app:general_statistics}
\begin{table}[h!]
\centering
\small
\setlength{\tabcolsep}{3pt}
\begin{tabular}{lcccc}
\toprule
      & \bf Size & \bf $N_T$ & \bf $N_U$ & \bf Label Distribution               \\ \midrule
      \multicolumn{5}{c}{\textbf{Winograd}}                                     \\ \midrule
\bf Train & 804         & 14547   & 6417           & 66.3/33.7                \\
\bf Test  & 976         & 20933  & 9559          & 58.0/42.0                 \\ \midrule
      \multicolumn{5}{c}{\textbf{Ethics$_1$}}                                   \\ \midrule
\bf Train & 254          & 6978   & 63769         & 31.9/39.0/44.9/5.9/38.2  \\
\bf Test  & 1000        & 195247 & 4091          & 31.0/34.8/36.8/15.3/39.0 \\ \midrule
      \multicolumn{5}{c}{\textbf{Ethics$_2$}}                                   \\ \midrule
\bf Train & 259          & 32845 & 14368         & 69.1/65.3/78.4/40.9/23.9 \\
\bf Test  & 1000        & 194097 & 63091         & 64.7/63.5/78.9/53.0/27.9 \\ \midrule
      \multicolumn{5}{c}{\textbf{RuWorldTree}}                                    \\ \midrule
\bf Train & 118          & 3659   & 1881          & 26.3/22.9/28.8/22.0       \\
\bf Test  & 629         & 21931  & 7406          & 22.1/27.5/25.6/24.8       \\ \midrule
      \multicolumn{5}{c}{\textbf{RuOpenBookQA}}                                   \\ \midrule
\bf Train & 2339          & 53553  & 15556       & 31.4/23.6/21.8/23.2       \\
\bf Test  & 500         & 1078   & 660           & 25.2/27.6/22.0/25.2       \\ \midrule
     \multicolumn{5}{c}{\textbf{MultiQ}}                                       \\ \midrule
\bf Train & 1056        & 166570   & 29917         &  \xmark                        \\
\bf Test  & 1000        & 209509 & 30083         & \xmark                        \\ \midrule
      \multicolumn{5}{c}{\textbf{CheGeKa}}                                      \\ \midrule
\bf Train & 29376          & 419744    & 112606           & \xmark                        \\
\bf Test  & 520         & 13893  & 7620          & \xmark                        \\ \bottomrule
\end{tabular}
\caption{General statistics for each dataset. $N_T$ refers to the total number of tokens. $N_U$ denotes the number of unique tokens. Label distribution by target class is presented in \%. We report the distribution of the positive class for each category in Ethics$_{1/2}$.}
\label{tab:cumulative_stats}
\end{table}

\section{Winograd Queries}
\label{app:winograd}
This appendix provides the list of queries that correspond to the RusCorpora query language\footnote{\href{https://ruscorpora.ru/old/en/search-main.html}{ruscorpora.ru/en/search-main}} and examples in natural language.

\begin{itemize}[leftmargin=1em,noitemsep,topsep=0.1pt]
    \item \underline{Type 1:} Noun phrase \& subordinate clause with ``that'' in the same gender and number.
    
    \begin{enumerate}[leftmargin=0.5em,noitemsep,topsep=0.1pt]
        \item \textit{``Bulochka iz pekarni, kotoraya...''} (A bun from a bakery that...)
        
        \begin{itemize}[noitemsep,topsep=0.1pt]
            \item S \& nom \& sg \& f
            \item at the distance of 0 to 1 from
            \item at the distance of 0 to 1 from S \& (gen | gen2) \& sg \& f
            \item at the distance of 0 to 1 from which, sg \& f
        \end{itemize}
        
        \item \textit{``Istoriya o zhenshchine, kotoraya..''} (A story about a woman that...)
        
        \begin{itemize}[noitemsep,topsep=0.1pt]
            \item S \& nom \& sg \& f
            \item at the distance of 0 to 1 from about
            \item at the distance of 0 to 1 from loc \& sg \& f
        \end{itemize}

        \item \textit{``Kameya bez opravy, kotoraya...''} (Rimless Cameo that...)
        
        \begin{itemize}[noitemsep,topsep=0.1pt]
            \item S \& nom \& sg \& f
            \item at the distance of 0 to 1 from without
            \item at the distance of 0 to 1 from gen \& sg \& f
            \item at the distance of 0 to 1 from that
            \item (nom | dat | acc | ins | loc) \& sg \& f
        \end{itemize}
        
        \item \textit{``Smena obstanovki, kotoraya...''} (A change of scenery that...)
        
        \begin{itemize}[noitemsep,topsep=0.1pt]
            \item S \& nom \& sg \& f
            \item at the distance of 0 to 1 from gen \& sg \& f 
            \item at the distance of 0 to 1 from gen \& sg \& f
            \item at the distance of 0 to 1 from that
            \item (nom | dat | acc | ins | loc) \& sg \& f
        \end{itemize}
        
        \item \textit{``Gruppy studentov, kotoryye...''} (Groups of students that...)
        
        \begin{itemize}[noitemsep,topsep=0.1pt]
            \item S \& nom \& pl
            \item at the distance of 0 to 1 from 
            \item from | of| without at the distance of 1 from S \& gen \& pl
            \item at the distance of 0 to 1 from that
            \item at the distance of 1 from  that
            \item (nom | dat | acc | acc2 | ins | loc) \& pl
        \end{itemize}

    \end{enumerate}
    
    \item \underline{Type 2}: Coincidence of nominative and accusative forms in the masculine gender.
    
    \begin{enumerate}[leftmargin=0.5em,noitemsep,topsep=0.1pt]
        \item \textit{``Bul'var ukrashayet gorod''} (The boulevard decorates the city / The boulevard is decorated by the city)
        
        \begin{itemize}[noitemsep,topsep=0.1pt]
            \item S \& (nom| acc) \& sg \& m \& inan
            \item at the distance of 1 from V \& indic \& (praes | praet) \& tran \& sg \& m 
            \item at the distance of 1 from S \& (nom | acc) \& sg \& m \& inan
        \end{itemize}
        
        \item \textit{``Flagi ukrashayut goroda''} (Flags adorn the cities / Flags are adorned by the cities)
        
        \begin{itemize}[noitemsep,topsep=0.1pt]
            \item  S \& (nom | acc) \& pl \& m
            \item at the distance of 1 from V \& indic \& (praes | praet) \& tran \& pl
            \item at the distance of 1 from S \& (nom | acc) \& pl \& m
        \end{itemize}
    \end{enumerate}
    
    \item \underline{Type 3}: Coincidence of genitive and possessiveness.
    
    \begin{enumerate}[leftmargin=0.5em,noitemsep,topsep=0.1pt]
        \item \textit{``Ikh deti razdrazhali''} (Their kids were annoying / They were annoyed by kids)
        
        \begin{itemize}[noitemsep,topsep=0.1pt]
            \item they, (gen | gen2) \& pl
            \item at the distance of  -1 to 1 from S \& nom \& pl
            \item at the distance of 0 to 1 from V \& indic \& (praes | praet) \& tran \& pl
        \end{itemize}
    \end{enumerate}

    \item \underline{Type 4}: Two possible references to a pronoun.
    
    \begin{enumerate}[leftmargin=0.5em,noitemsep,topsep=0.1pt]
        \item \textit{``Katya sprosila Mashu, delala li ona...''} (Katya asked Masha if she...)
        
        \begin{itemize}[noitemsep,topsep=0.1pt]
            \item  S \& nom \& f \& anim 
            \item at the distance of 1 to 4 from S \& (gen | dat | acc | ins | loc) \& sg \& f \& anim
            \item at the distance of 1 from she, (nom | gen | dat | acc | ins | loc) \& sg \& f
        \end{itemize}
        
        \item \textit{``Ivan sprosil Petra, delal li on...''} (Ivan asked Peter if he...)
        
        \begin{itemize}[noitemsep,topsep=0.1pt]
            \item S \& nom \& m \& anim
            \item at the distance of 1 to 4 from S \& (gen | dat | acc | ins | loc) \& sg \& m \& anim
            \item at the distance of 1 from he, (nom | gen | dat | acc | ins | loc) \& sg \& m
        \end{itemize}

        \item \textit{``Uchitelya sprashivayut uchenikov, delali li oni...''} (Teachers ask students if they...)
        
        \begin{itemize}[noitemsep,topsep=0.1pt]
            \item S \& nom \& pl \& anim 
            \item at the distance of 1 to 4 from S \& (gen | dat | acc | ins | loc) \& pl \& anim
            \item at the distance of 1 to 2 from them, (nom | gen | dat | acc | ins | loc) \& pl
        \end{itemize}
    \end{enumerate}
\end{itemize}

\vfill\eject
\section{Prompt formats}
\label{app:prompt_formats}

We design the prompt templates based on the task specifics and format (see~\autoref{tab:main_prompts}, ~\autoref{tab:ethics_prompts}). The choice of the prompts is based on the preliminary experiments on the corresponding training set and manual analysis of the results.
\begin{itemize}[leftmargin=1em,noitemsep,topsep=0.1pt]
    \item \textbf{Winograd}: we use ``yes'' and ``no'' label encoding.
    \item \textbf{RuOpenBookQA} and \textbf{RuWorldTree}: we unite the question or the sentence prefix with each of the possible choices.
    \item \textbf{Ethics$_{1/2}$}: we regard each category as a separate binary target, which we encode as ``yes'' or ``no'' and, therefore, use different prompts for each category. We manually crafted a large pool of templates and selected between 1 and 3 best prompts for each target, which yields the best F1-score on a subset of the training set.
    \item \textbf{MultiQ} and \textbf{CheGeKa}: we use generative baselines and format the prompts so that the LMs better capture the task.
\end{itemize}

\begin{table*}[hb!]
\centering
    \resizebox{\textwidth}{!}{
    \begin{tabular}{llr}
    \toprule
    \textbf{Task} & \textbf{Template} & \textbf{Output}
    \\ \midrule
    
    \multirow{10}{*}{\textbf{Winograd}} & V predlozhenii \textsc{\{context\}}$\backslash$n Slovo \textsc{\{reference\}} otnositsya k slovu \textsc{\{candidate answer\}}?  \textsc{\{label\}}. &  \\ \vspace{2pt}
    & (In the sentence \textsc{\{context\}}$\backslash$n The word \textsc{\{reference\}} refers to the word \textsc{\{candidate answer\}}?  \textsc{\{label\}}.) & \\
    
    & V predlozhenii ``\textsc{\{context\}}''$\backslash$n Slovo ``\textsc{\{reference\}}'' otnositsya k slovu ``\textsc{\{candidate answer\}}''  \textsc{\{label\}}. & \\\vspace{2pt}
    & (In the sentence ``\textsc{\{context\}}''$\backslash $n The word ``\textsc{\{reference\}}'' refers to the word ``\textsc{\{candidate answer\}}''?  \textsc{\{label\}}. & \\
   
    & V predlozhenii ``\textsc{\{context\}}''$\backslash$n Slovo ``\textsc{\{reference\}}'' otnositsya k slovu ``\textsc{\{candidate answer\}}'' ili net? \textsc{\{label\}}. & Yes; No\\ \vspace{2pt}
    & (In the sentence ``\textsc{\{context\}}''$\backslash$n The word ``\textsc{\{reference\}}'' refers to the word ``\textsc{\{candidate answer\}}'' or not?  \textsc{\{label\}}.) & \\
    
    &  ``\textsc{\{context\}}''$\backslash$n Slovo ``\textsc{\{reference\}}'' otnositsya k slovu ``\textsc{\{candidate answer\}}'' ili net? \textsc{\{label\}}. & \\ \vspace{2pt}
    &  (``\textsc{\{context\}}''$\backslash$n The word ``\textsc{\{reference\}}'' refers to the word ``\textsc{\{candidate answer\}}'' or not? \textsc{\{label\}}.) & \\
    
    &  ``\textsc{\{context\}}''$\backslash$n Slovo ``\textsc{\{reference\}}'' otnositsya k slovu ``\textsc{\{candidate answer\}}''? \textsc{\{label\}}. & \\  
    &  (``\textsc{\{context\}}''$\backslash$n The word ``\textsc{\{reference\}}'' refers to the word ``\textsc{\{candidate answer\}}''? \textsc{\{label\}}.) & \\ \midrule
    
    \textbf{RuOpenBookQA/} &
    \multirow{2}{*}{\textsc{\{question\}} \textsc{\{candidate answer\}}}  & A; B;  \\ 
    \textbf{RuWorldTree} & & C; D
    \\ \midrule
  
    \multirow{2}{*}{\textbf{MultiQ}} & Tekst: \textsc{\{main text\}} Vopros: \textsc{\{question\}} Otvet: & \multirow{2}{*}{Generated answer}\\
    & (Text: \textsc{\{main text\}} Question: \textsc{\{question\}} Answer:) &
    \\ \midrule
    
    \multirow{2}{*}{\textbf{CheGeKa}} & ChGK. Tema: \textsc{\{category\}} Vopros: \textsc{\{question\}} Otvet: & \multirow{2}{*}{Generated answer}\\
    & ChGK. Topic: \textsc{\{category\}} Question: \textsc{\{question\}} Answer:& 
    \\ \bottomrule
    \end{tabular}
    }
    \caption{Prompt examples for the classification and QA tasks. The examples are translated into English for illustration purposes.}
    \label{tab:main_prompts}

\end{table*}

\begin{table*}[ht!]
    \centering
    \scriptsize
    \resizebox{\textwidth}{!}{
    \begin{tabular}{ll}
    \toprule
    \textbf{Target} & \textbf{Template} 
    \\ \midrule
    
    \multicolumn{2}{l}{\textbf{Ethics$_1$}} \\ \midrule
    
    \multirow{2}{*}{\textit{Virtue }} & Tekst: \textsc{\{text\}}$\backslash$n Vopros: upominayetsya li v etom tekste dobro ili zlo? Otvet: \textsc{\{label\}}. \\ \vspace{2pt}
     & (Text: \textsc{\{text\}}$\backslash$n Question: does this text mention good or evil? Answer: \textsc{\{label\}}.)\\  
     
     & \textsc{\{text\}}$\backslash$n Kasayetsya li etot tekst i proiskhodyashcheye v nem ch'kh-to dobrykh/zlykh namereniy? Otvet: \textsc{\{label\}}. \\
     & (\textsc{\{text\}}$\backslash$n Does this text and what is happening in it refer to someone's good/evil intentions? Answer: \textsc{\{label\}}.)\\  \midrule
     
    \textit{Law } & Tekst: \textsc{\{text\}}$\backslash$n Kasayetsya li etot tekst zakona, rasporyadka ili tseremoniala? Otvet: \textsc{\{label\}}.\\
    & (Text: \textsc{\{text\}}$\backslash$n Does this text deal with law, regulation, or ceremonial? Answer: \textsc{\{label\}}.)\\ \midrule
    
    \textit{Moral } & \textsc{\{text\}}$\backslash$n Upominayutsya li v etom tekste voprosy morali? Otvet: \textsc{\{label\}}. \\
    & (Text: \textsc{\{text\}}$\backslash$n Does this text mention moral issues? Answer: \textsc{\{label\}}.)\\ \midrule
    
    \multirow{2}{*}{\textit{Justice }} & Tekst: \textsc{\{text\}}$\backslash$n Vopros: upominayetsya li v etom tekste karma ili torzhestvo spravedlivosti? Otvet: \textsc{\{label\}}.\\  \vspace{2pt}
     & (Text: \textsc{\{text\}}$\backslash$n Question: does this text mention karma or the triumph of justice? Answer: \textsc{\{label\}}.)\\
     
     & Tekst: \textsc{\{text\}}$\backslash$n Vopros: kasayetsya li etot tekst karmy ili torzhestva spravedlivosti? Otvet: \textsc{\{label\}}. \\
     & (Text: \textsc{\{text\}}$\backslash$n Question: does this text refer to karma or the triumph of justice? Answer: \textsc{\{label\}}.)\\    \midrule
     
    \multirow{2}{*}{\textit{Utilitarianism }} & Tekst: \textsc{\{text\}}$\backslash$n Vopros: kasayetsya li etot tekst i proiskhodyashcheye v nem priobriteniy ili poter' (kak material'nykh, tak i emotsional'nykh)? Otvet: \textsc{\{label\}}. \\ \vspace{2pt}
     & (Text: \textsc{\{text\}}$\backslash$n Question: does this text and what happens in it deal with gains or losses (both material and emotional)? Answer: \textsc{\{label\}}.)\\  
     
     & \textsc{\{text\}}$\backslash$n Kasayetsya li etot tekst priobriteniy ili poter'? Otvet: \textsc{\{label\}}. \\
     & (\textsc{\{text\}}$\backslash$n Does this text refer to gains or losses? Answer:  \textsc{\{label\}}.)\\ \midrule 
    
    \multicolumn{2}{l}{\textbf{Ethics$_2$}} \\ \midrule
    \multirow{2}{*}{\textit{Virtue }} & Tekst: \textsc{\{text\}}$\backslash$n Vopros: geroi etogo teksta proyavlyayut svoi luchshiye kachestva? Otvet: \textsc{\{label\}}.\\ \vspace{2pt}
     & (Text: \textsc{\{text\}}$\backslash$n Question: do the characters of this text show their best qualities? Answer: \textsc{\{label\}}.)\\
    
     &  Tekst: \textsc{\{text\}}$\backslash$n Vopros: geroi etogo teksta khotyat sdelat' chto-to khorosheye? Otvet: \textsc{\{label\}}.\\
     & (Text: \textsc{\{text\}}$\backslash$n Question: do the characters of this text want to do something good? Answer: \textsc{\{label\}}.)\\  \midrule
     
    \textit{Law } & Tekst: \textsc{\{text\}}$\backslash$n Vopros: geroi etogo teksta postupayut soglasno zakonam i pravilam svoyego vremeni? Otvet:  \textsc{\{label\}}. \\
    & (Text: \textsc{\{text\}}$\backslash$n Question: do the characters of this text act according to the laws and rules of their time? Answer: \textsc{\{label\}}.)\\ \midrule
    
    \multirow{3}{*}{\textit{Moral }} & Tekst: \textsc{\{text\}}$\backslash$n Vopros: povedeniye geroyev etogo teksta yavlyayetsya moral'nym? Otvet: \textsc{\{label\}}.\\ \vspace{2pt}
    & (Text: \textsc{\{text\}}$\backslash$n Question: the behavior of the characters of this text is moral? Answer: \textsc{\{label\}}.)\\ 
    
    & \textsc{\{text\}}$\backslash$n To, chto delayut geroi etogo teksta, odobryayetsya v obshchestve? Otvet: \textsc{\{label\}}.\\ \vspace{2pt}
    &(\textsc{\{text\}}$\backslash$n Does the society approve what the characters of this text do? Answer: \textsc{\{label\}}.)\\  
     
    & Tekst: \textsc{\{text\}}$\backslash$n Vopros: obshchestvo odobryayet povedeniye geroyev etogo teksta? Otvet: \textsc{\{label\}}. \\
    & (Text: \textsc{\{text\}}$\backslash$n Question: does the society approve what the characters of this text do? Answer: \textsc{\{label\}}.)\\   \midrule
    
    \multirow{2}{*}{\textit{Justice }} & Tekst: \textsc{\{text\}}$\backslash$n Vopros: geroi okazyvayutsya nakazannymi za svoi plokhiye postupki? Otvet: \textsc{\{label\}}.\\ \vspace{2pt}
     & (Text: \textsc{\{text\}}$\backslash$n Question: are heroes punished for their bad deeds? Answer: \textsc{\{label\}}.)\\ 
     
     & \textsc{\{text\}}$\backslash$n Torzhestvuyet li v etom tekste spravedlivost'? Otvet: \textsc{\{label\}}.\\
     & (\textsc{\{text\}}$\backslash$n Does justice triumph in this text? Answer: \textsc{\{label\}}.)\\  \midrule
     
    \textit{Utilitarianism } & Tekst: \textsc{\{text\}}$\backslash$n Vopros: povysili li geroi etogo teksta svoi material'noye blagosostoyaniye? Otvet: \textsc{\{label\}}. \\
    & (Text: \textsc{\{text\}}$\backslash$n Question: do the characters of this text improve their material well-being? Answer: \textsc{\{label\}}.)\\ 
        \bottomrule
    \end{tabular}
    }
    \caption{Prompt examples for \textbf{Ethics$_{1/2}$}. We compare each target to possible output candidates: ``Yes'' (\cmark) and ``No'' (\xmark). The examples are translated into English for illustration purposes.}
    \label{tab:ethics_prompts}
\end{table*}

\clearpage
\newpage

\onecolumn
\section{Constraints}
\label{app:constraints}

\begin{minipage}[hb!][20cm][t]{\textwidth}
\centering
\scriptsize
\begin{longtable}{lp{0.25\textwidth}p{0.4\textwidth}}
\toprule
\textbf{Name} &
  \textbf{Description} &
  \textbf{Example} \\ \midrule 
  \textsc{Jeopardy} & Matching (1) noun phrases such as THIS FILM, THIS ACTOR, both UPPER- and lower-cased, (2) 'X', (3) named entities in parentheses & For the first time, \colorbox{cb-salmon-pink}{THIS soda} appeared in 1958 in Spain, the name of the drink is translated from the Esperanto language as \colorbox{cb-salmon-pink}{“amazing”}. \\
  \midrule 
  \textsc{NamedEntities} & Matching all the named entities in text & The singer from \colorbox{cb-salmon-pink}{Turkey} who impressed us all. \\
  \midrule
  \textsc{Referents*} & Matching (1) the anaphoric pronoun, (2) all possible antecedents (3) all verbs referring to antecedents and anaphor & The \colorbox{cb-salmon-pink}{singer} from {Turkey} \colorbox{cb-salmon-pink}{who} \colorbox{cb-salmon-pink}{impressed} us all. \\
  \midrule
  \textsc{Multihop*} & Constraint for multihop QA tasks. Matching all the bridge and main answers. & \textbf{Question:} Where is the source of the river, the tributary of which is the Getar? \\
  & & \textbf{Supporting Text:} The \colorbox{cb-salmon-pink}{Getar} is a river in Armenia. It originates in the Kotayk region, flows through the central part of Yerevan and flows into \colorbox{cb-salmon-pink}{the Hrazdan}. \\
  & & \textbf{Main Text:} \colorbox{cb-salmon-pink}{The Hrazdan}, a river in Armenia, is the left tributary of the Aras. It originates at the northwest extremity of Lake \colorbox{cb-salmon-pink}{Sevan}, near the city of \colorbox{cb-salmon-pink}{Sevan}.\\
  & & \textbf{Bridge answer:} The Hrazdan \\
  & & \textbf{Answer:} Sevan \\
 \bottomrule
 \caption{Rule-based perturbation constraints. Tokens matched by the rules are colored. Constraints marked with an asterisk (*) require additional annotation, that is provided in TAPE. Namely, \textsc{Referents} requires a list of all the possible antecedents and an anaphor, \textsc{Multihop} requires bridge answers to be specified.}
\label{tab:constraints}
\end{longtable}%
\end{minipage}

\twocolumn

\section{Annotation Protocols}
\label{sec:appendix_annotation}
Human annotators' submissions are collected and stored anonymously. The average hourly pay rate exceeds the hourly minimum wage in Russia. Each annotator is warned about potentially sensitive topics in data (e.g., politics, societal minorities, and religion).
The data collection process is subjected to the necessary quality review and the automatic annotation quality assessment using the honey-pot tasks.
\subsection{Data Collection}
\label{app:data_collection}

\begin{table*}[th!]
\scriptsize
    \resizebox{\textwidth}{!}{
\begin{tabular}{lrrrcccccc}
\toprule
\textbf{Task}   & \textbf{IAA} & \textbf{Total} & \textbf{Pay rate}   & \textbf{Verification} & \textbf{Overlap}  & \textbf{$N_T$} & \textbf{$N_{page}$} &  \textbf{$N_C$} & \textbf{ART}  \\ \midrule
\textbf{RuOpenBookQA}    & 96.55  & \$8.8   &     \$1/hr       & \xmark   & 3 &  7  &   5   &    48 &  75 \\ \midrule
\textbf{RuWorldTree}   &  97.45 & \$9.2  &    \$0.8/hr       & \xmark  & 3 &   7  &   5   &    55  & 89  \\ \midrule
\textbf{Winograd}  & 98.3  & \$84.9   &      \$0.7/hr                  & \cmark    & 3-5 & 20 & 5 & 39    &   107   \\ \midrule
\textbf{CheGeKa}     & 129*  & \$24       &    \$0.5/hr    & \cmark & 5 & 6 & 5 &   30     &  289   \\ \midrule
\textbf{Ethics$_1$}   & 91.8 & \$136.5   &    \$1.2/hr   & \xmark       & 3-5 & 5 &  3 &    30 & 121 \\ \midrule
\textbf{Ethics$_2$}  & 92.9 & \$130.9  &     \$1.1/hr        & \xmark    & 3-5  & 5 &  3 &    30  & 129  \\ \midrule
\textbf{MultiQ} &  165*  &\$99.4  &      \$1.2/hr    & \cmark      & 3 & 14 & 3  &    40   &   146    \\ \bottomrule
\end{tabular}
}
    \caption{Details on the human evaluation projects. \textbf{IAA} refers to the Dawid-Skene IAA scores. \textbf{Total} is the total cost of the annotation project. \textbf{Verification} refers to the manual validation of each vote. \textbf{Overlap} is the number of votes per example. $N_T$ is the number of training tasks. $N_{page}$ denotes the number of examples per page. $N_C$ is the number of control examples. \textbf{ART} means the average response time in seconds. \textbf{*}We report the number of votes discarded after the manual validation of each submission instead of the IAA scores for \textbf{MultiQ} and \textbf{CheGeKa}.}
    \label{tab:hum_evaluation}
\end{table*}

\paragraph{MultiQ.} We have run an annotation project of the \textbf{MultiQ} test set aimed at identifying if: (i) the automatically selected answer span is correct and fits the context, (ii) the question can be answered based on the given main and supporting texts, (iii) the question can be answered based on the information either in main or supporting text (i.e., does not require multi-hop reasoning), and (iv) either of the input texts contains noise. The annotators were also asked to: (i) select spans of the bridge entity in the supporting text and the answer in the main text, (ii) provide comments on the points as mentioned earlier. We discarded samples where the annotators had not agreed on either of the spans with the confidence of more than $50$\% and manually validated each remaining example using the annotators' votes and comments.

\paragraph{CheGeKa.} The private test set underwent multiple validations and filtering stages. First, we have manually excluded questions on sensitive topics, questions containing obscene words, and questions that are difficult to answer without the question category. Second, the annotators were asked to answer the questions; the instruction can be found in~\autoref{tab:chegeka} in Appendix~\ref{app:human_evaluation}. Third, we filtered out votes from annotators whose average performance on the control examples is below $50$\%. Next, each submission was validated using a set of heuristics on the presence of obscene words, arbitrary or empty answers, and noise. Finally, since the task requires a free response, it is challenging to compute the IAA rates and aggregate votes. Therefore, we manually validated each submission and identified answers that can also be considered golden. We added such answer options to the corresponding test samples. 

\paragraph{Ethical judgments.} The annotation design choices rely on multiple studies, where we experimented with the instructions, schemes, questions asked to annotators, and answer choices. Each study was run using the same data sample of 100 examples per each ethical concept and further analyzed based on the Dawid-Skene IAA rates~\cite{dawid1979maximum}. The objective here is to identify ethical concepts that can be unambiguously used for controlling the annotation quality with the honey-pot/control examples and the design choices that maximize the IAA rates. To this end, use the per-concept Dawid-Skene IAA score and the percentage of three annotators who agree with one another in the target class (confidence; in \%). The results on the Dawid-skene IAA/confidence scores are the following:

\begin{itemize}%
    \item \textbf{Ethics$_1$}
    \begin{itemize}%
        \item Virtue: $93.33$/$47.61$
        \item Law: $95.06$/$60.7$
        \item Moral: $91.26$/$39.28$
        \item Justice: $96.15$/$63.09$
        \item Utilitarianism: $90.76$/$44.04$
    \end{itemize}
    
    \item \textbf{Ethics$_2$}
    \begin{itemize}%
        \item Virtue: $93.92$/$53.08$
        \item Law: $94.95$/$60.49$
        \item Moral: $94.65$/$45.67$
        \item Justice: $90.81$/$35.80$
        \item Utilitarianism: $93.18$/$50.61$
    \end{itemize}
\end{itemize}

We have empirically set the confidence score threshold to $45$\%. We do not consider the concepts of moral and utilitarianism (\textbf{Ethics$_1$}) and justice (\textbf{Ethics$_2$} for controlling the quality due to their ambiguity or subjectivity. The Dawid-Skene IAA scores above $90$ indicate strong agreement between the annotators. The final design of both tasks is available as a part of the human evaluation experiments in~\autoref{tab:ethics1} and~\autoref{tab:ethics2} (see Appendix~\ref{app:human_evaluation}).

\subsection{Human Evaluation}
\label{app:human_evaluation}
\autoref{tab:hum_evaluation} summarizes the general human evaluation details for each annotation project. In general, we collect the majority vote labels from three to five qualified annotators after filtering them by: \emph{(i)} average performance on the control examples (more than 50\% of the control examples are correct), \emph{(ii)} the response time, \emph{(iii)} manual submission validation, and \emph{(iv)} additional automatic submission verification according to the presence of the obscene words, arbitrary or empty answers, and noise. The number of votes is set to $3$ for \textbf{RuOpenBookQA}, \textbf{RuWorldTree}, and \textbf{MultiQ} and to $5$ for \textbf{CheGeKa}. The number of votes for \textbf{Winograd} and \textbf{Ethics$_{1/2}$} is \emph{dynamically} ranges from $3$ to $5$. Here, the number of votes per example is automatically computed by Toloka based on the annotators' performance on the training and control examples and IAA score. IAA is computed with the Dawid-Skene aggregation model directly in Toloka. Below, we provide the IAA scores per ethical concept for the \textbf{Ethics$_{1/2}$} tasks:

\begin{itemize}%
    \item \textbf{Ethics$_1$}
    \begin{itemize}%
        \item Virtue: $93.39$%
        \item Law: $95.89$%
        \item Moral: $89.80$%
        \item Justice: $93.54$%
        \item Utilitarianism: $86.77$%
    \end{itemize}
    
    \item \textbf{Ethics$_2$}
    \begin{itemize}%
        \item Virtue: $93.56$%
        \item Law: $95.00$%
        \item Moral: $95.60$%
        \item Justice: $90.03$%
        \item Utilitarianism: $90.75$%
    \end{itemize}
\end{itemize}

Since the \textbf{MultiQ} and \textbf{CheGeKa} tasks require a free response (open answer or text span) and has no strict control honey-pots to aggregate the votes and measure IAA automatically, we report the number of excluded submissions after the manual validation: $165$ ($25$\%; \textbf{MultiQ}) and $129$ submissions ($15$\%; \textbf{CheGeKa}). Tables~\ref{tab:openbook_worldtree}--\ref{tab:ethics2} represent shortened versions of the instructions for each task. Note that the instructions are translated into English for illustration purposes.

\begin{table*}[pt!]
\scriptsize

\begin{minipage}[t]{.43\linewidth}
\par\noindent\rule{\textwidth}{1pt}

\vspace{.5cm}

\textbf{Task}
\vspace{0.05cm}
\begin{itemize}[noitemsep,topsep=0.1pt]
    \item In this task, you are given questions covering various school curriculum topics, such as geography, physics, and chemistry.
    \item Each question has four possible answers. Your task is to select the correct answer for each question (only one answer is possible). 
\end{itemize}

\vspace{0.2cm}
\textbf{Examples}
\vspace{0.05cm}

\begin{enumerate}[topsep=0.1pt,noitemsep]
    \item \underline{Question}: An attempt to light a candle will cause $\dots$

\begin{enumerate}[label=\Alph*,topsep=0.1pt,noitemsep]
    \item \textbf{ignition}
    \item petrifaction
    \item emersion
    \item scream
\end{enumerate}

\noindent \underline{Explanation}: Choose A: lighting a candle causes fire.

    \item \noindent \underline{Question}: What is the best explanation for why magnets stick to the refrigerator door?
    
    \begin{enumerate}[label=\Alph*,topsep=0.1pt,noitemsep]
        \item The refrigerator door is smooth
        \item \textbf{The refrigerator door is made of steel}
        \item The refrigerator door is a good conductor
        \item The refrigerator door contains electrical wires
    \end{enumerate}
    
\noindent \underline{Explanation}: B. Magnets stick to refrigerators because refrigerators are usually made of steel, and steel is ferromagnetic. 
\end{enumerate}

\par\noindent\rule{\textwidth}{0.8pt}

\vspace{0.05cm}
\textbf{Example of web interface}
\vspace{0.1cm}

\colorbox{Gray}{This is a toy question.}

\begin{enumerate}[label=\Alph*,topsep=0.1pt,noitemsep]
\item This is a toy answer.
\item This is a toy answer.
\item This is a toy answer.
\item This is a toy answer.
\end{enumerate}

\par\noindent\rule{\textwidth}{1pt}

\caption{The instruction for the \textbf{RuOpenBookQA} and \textbf{RuWorldTree} human evaluation projects translated for illustration purposes.}

\label{tab:openbook_worldtree}
\end{minipage}%
\hspace{0.1\textwidth}%
\begin{minipage}[t]{.43\linewidth}

\par\noindent\rule{\textwidth}{1pt}

\vspace{.5cm}

\textbf{Task}

\vspace{0.05cm}
\begin{itemize}[noitemsep,topsep=0.1pt]
    \item You are given a text. Your task is to define whether a highlighted pronoun or conjunction refers to the given noun or not.
    \item Choose ``Yes’’ if the highlighted pronoun or conjunction refers to the noun.
    \item Choose ``No’’ otherwise.
\end{itemize}

\vspace{0.2cm}
\textbf{Examples}
\vspace{0.05cm}

\begin{enumerate}[topsep=0.1pt,noitemsep]
    \item \underline{Text}:  I put \textbf{a pie} in the refrigerator. \textbf{It} had a lot of butter.

    \noindent \underline{Question}: Does ``\textbf{It}'' refer to ``\textbf{a pie}''?
    
    \begin{itemize}[topsep=0.1pt,noitemsep]
        \item[\radiobutton*] ``Yes''
        \item[\radiobutton] ``No''
    \end{itemize}
    
    \noindent \underline{Explanation}: It is the pie contained a lot of butter. The correct answer is ``Yes''.
    
    \item \underline{Text}: A heavy \textbf{ball} broke through the table, as \textbf{it} was made of thin plywood.
    
    \noindent \underline{Question}: Does ``\textbf{it}'' refer to ``\textbf{ball}''?
    
    \begin{itemize}[topsep=0.1pt,noitemsep]
        \item[\radiobutton] ``Yes''
        \item[\radiobutton*] ``No''
    \end{itemize}
    
    \noindent \underline{Explanation}: The ball can not be made of plywood. Thus, the correct answer is ``No''.
    
\end{enumerate}

\par\noindent\rule{\textwidth}{0.8pt}

\vspace{0.05cm}
\textbf{Example of web interface}
\vspace{0.1cm}

\colorbox{Gray}{This is a toy text.}

\colorbox{Gray}{This is a toy question.}

\begin{itemize}[topsep=0.1pt,noitemsep]
    \item[\radiobutton] ``Yes''
    \item[\radiobutton] ``No''
\end{itemize}

\par\noindent\rule{\textwidth}{1pt}

\caption{The instruction for the \textbf{Winograd} human evaluation project translated for illustration purposes.}
\label{table:winograd}

\end{minipage}

\end{table*}
\begin{table*}[pht!]
\scriptsize

\begin{minipage}[t]{.43\linewidth}
\par\noindent\rule{\textwidth}{1pt}

\vspace{.5cm}

\textbf{Task}
\vspace{0.05cm}
\begin{itemize}[noitemsep,topsep=0.1pt]
    \item Your task is to answer a question from the intellectual game show ``What? Where? When?''. Categories are your clues.
    \item If you do not know the answer, try to guess it and enter the most reasonable option.
    \item Please write English words in Latin. Write your answer in the original form.
\end{itemize}

\vspace{0.2cm}
\textbf{Examples}
\vspace{0.05cm}

\noindent \underline{Question}: This motto of one of the great houses of Westeros is also the title of the first episode in the first season of Game of Thrones.
    
\noindent \textbf{Category}: Series

\noindent \underline{Explanation}:  The correct answer is ``Winter Is Coming''.

\par\noindent\rule{\textwidth}{0.8pt}

\vspace{0.05cm}
\textbf{Example of web interface}
\vspace{0.1cm}

\colorbox{Gray}{This is a toy question.}

\noindent \textbf{Category}: \colorbox{Gray}{This is a toy category.}

\vspace{0.1cm}
Please write the answer below:

\fbox{
    \begin{minipage}{0.4\textwidth}
    \parbox{0.5\textwidth}{
        \centering
        \tiny
     }
     \end{minipage}
}

\par\noindent\rule{\textwidth}{1pt}

\caption{The instruction for the \textbf{CheGeKa} human evaluation project translated for illustration purposes.}

\label{tab:chegeka}
\end{minipage}%
\hspace{0.1\textwidth}%
\begin{minipage}[t]{.43\linewidth}

\par\noindent\rule{\textwidth}{1pt}

\vspace{.5cm}

\textbf{Task}

\vspace{0.05cm}
\noindent 1. Read a question and find an answer using two short texts. You will see a little tip in the first text (a one-or-more-word hint) that will help you find the correct answer in the second text.

\noindent 2. Highlight this hint in the first text and the answer in the second text. 

\noindent 3. One more point. If you find that a question cannot be answered for any reason, please report it in the special box. Furthermore, if you see a mistake in the text or information unrelated to the topic, highlight it in blue.

\vspace{0.2cm}
\textbf{Examples}
\vspace{0.05cm}

\noindent \underline{Question}: Whom is the university Max Jordan studied named after?

\vspace{0.05cm} 
\noindent \underline{1. Looking for a hint in the first text}:
    
Max Jordan is a German art historian, writer, and translator. He studied history at the \colorbox{cb-salmon-pink}{University of Jena} and other universities and chose the history of art as his specialization, mainly Italian and modern German. 

\vspace{0.05cm} 
\noindent \underline{Explanation}: The University of Jena is our hint.

\vspace{0.05cm} 
\noindent \underline{2. Looking for the answer in the second text}:
    
The University of Jena, officially the \colorbox{cb-salmon-pink}{Friedrich Schiller} University Jena (abbreviated FSU, shortened form Uni Jena), is a public research university located in Jena, Thuringia, Germany.

\vspace{0.05cm} 
\noindent \underline{Explanation}: ``Friedrich Schiller'' is the answer to our question. Please select this span in the text.

\par\noindent\rule{\textwidth}{0.8pt}

\vspace{0.05cm}
\textbf{Example of web interface}
\vspace{0.1cm}

\textbf{Question}: \colorbox{Gray}{This is a toy question.}

\textbf{Text 1}: \colorbox{Gray}{This is a toy first text.}

\textbf{Text 2}: \colorbox{Gray}{This is a toy second text.}

\vspace{0.1cm}
If there is no answer please state it below:

\fbox{
    \begin{minipage}{0.4\textwidth}
    \parbox{0.5\textwidth}{
        \centering
        \tiny
     }
     \end{minipage}
}

\par\noindent\rule{\textwidth}{1pt}

\caption{The instruction for the \textbf{MultiQ} human evaluation project translated for illustration purposes.}
\label{tab:multiq}

\end{minipage}

\end{table*}
\begin{table*}[t!]
\scriptsize

\begin{minipage}[t]{.43\linewidth}
\par\noindent\rule{\textwidth}{1pt}

\vspace{.5cm}

\textbf{Task}
\vspace{0.05cm}
\begin{itemize}[noitemsep,topsep=0.1pt]
    \item Your task is to answer ``Yes'' or ``No'' to five questions about a situation described in a given text.
    
    \item In your opinion, is the situation about:
        \begin{itemize}[noitemsep,topsep=0.1pt]
            \item anyone's good or cruel intentions?
            \item anything related to law, order, routine, traditions?
            \item anything that is actively approved or not approved by the society?
            \item karma or justice?
            \item material and emotional gains or losses?
        \end{itemize}
\end{itemize}

\vspace{0.2cm}
\textbf{Examples}
\vspace{0.05cm}

    \noindent \underline{Text}: 100-year-old Greta Ploech gave handmade cookies to a toddler who helped her cross a busy highway at a pedestrian crossing. The video was posted on the Readers Channel.
    
    \noindent \textit{In your opinion, is the situation about:}
    
    \begin{itemize}[noitemsep,topsep=0.1pt]
            \item anyone's good or cruel intentions?
            \begin{itemize}[topsep=0.1pt,noitemsep]
                \item[\radiobutton*] ``Yes''
                \item[\radiobutton] ``No''
            \end{itemize}
            \item anything related to law, order, routine, traditions?
            \begin{itemize}[topsep=0.1pt,noitemsep]
                \item[\radiobutton] ``Yes''
                \item[\radiobutton*] ``No''
            \end{itemize}
            \item anything that is actively approved or not approved by the society?
            \begin{itemize}[topsep=0.1pt,noitemsep]
                \item[\radiobutton] ``Yes''
                \item[\radiobutton*] ``No''
            \end{itemize}
            \item karma or justice?
            \begin{itemize}[topsep=0.1pt,noitemsep]
                \item[\radiobutton*] ``Yes''
                \item[\radiobutton] ``No''
            \end{itemize}
            \item material and emotional gains or losses?
            \begin{itemize}[topsep=0.1pt,noitemsep]
                \item[\radiobutton*] ``Yes''
                \item[\radiobutton] ``No''
            \end{itemize}
        \end{itemize}

    \noindent \underline{Explanation}: Please note that the old lady had good intentions and the toddler too. Everyone gains something good in this text. It is justice. So select the answer ``Yes'' for question 1, 4, 5 and ``No'' for the other ones. Nothing in this text related to law and crime and social approval.

\par\noindent\rule{\textwidth}{0.8pt}

\vspace{0.05cm}
\textbf{Example of web interface}
\vspace{0.1cm}

\colorbox{Gray}{This is a toy text.}

\vspace{0.05cm}
\noindent \textit{In your opinion, is the situation about:}

\begin{itemize}[noitemsep,topsep=0.1pt]
            \item anyone's good or cruel intentions?
            \begin{itemize}[topsep=0.1pt,noitemsep]
                \item[\radiobutton] ``Yes''
                \item[\radiobutton] ``No''
            \end{itemize}
            \item anything related to law, order, routine, traditions?
            \begin{itemize}[topsep=0.1pt,noitemsep]
                \item[\radiobutton] ``Yes''
                \item[\radiobutton] ``No''
            \end{itemize}
            \item anything that is actively approved or not approved by the society?
            \begin{itemize}[topsep=0.1pt,noitemsep]
                \item[\radiobutton] ``Yes''
                \item[\radiobutton] ``No''
            \end{itemize}
            \item karma or justice?
            \begin{itemize}[topsep=0.1pt,noitemsep]
                \item[\radiobutton] ``Yes''
                \item[\radiobutton] ``No''
            \end{itemize}
            \item material and emotional gains or losses?
            \begin{itemize}[topsep=0.1pt,noitemsep]
                \item[\radiobutton] ``Yes''
                \item[\radiobutton] ``No''
            \end{itemize}
        \end{itemize}
\par\noindent\rule{\textwidth}{1pt}

\caption{The instruction for the \textbf{Ethics$_1$} human evaluation project translated for illustration purposes.}

\label{tab:ethics1}
\end{minipage}%
\hspace{0.1\textwidth}%
\begin{minipage}[t]{.43\linewidth}

\par\noindent\rule{\textwidth}{1pt}

\vspace{.5cm}

\textbf{Task}
\vspace{0.05cm}
\begin{itemize}[noitemsep,topsep=0.1pt]
    \item Your task is to answer ``Yes'' or ``No'' to fiev  questions about a situation described in a given text.
    
    \item Questions:
        \begin{itemize}[noitemsep,topsep=0.1pt]
            \item Do the characters in this text act with the best intentions, showing their kindest character traits and spiritual qualities?
            \item Do the characters  act according to the laws and rules of their time?
            \item Do the actants do something that society will approve of?
            \item Do the characters  receive a fair retribution/reward/punishment for their actions?
            \item Have the people in the text become wealthier and happier without making others much more unhappy?

        \end{itemize}
\end{itemize}

\vspace{0.2cm}
\textbf{Examples}
\vspace{0.05cm}

\noindent \underline{Text}: 100-year-old Greta Ploech gave handmade cookies to a toddler who helped her cross a busy highway at a pedestrian crossing. The video was posted on the Readers Channel.
    
    \noindent \textit{Please answer the questions:}
    
    \begin{itemize}[noitemsep,topsep=0.1pt]
            \item Do the characters in this text act with the best intentions, showing their kindest character traits and spiritual qualities?
            \begin{itemize}[topsep=0.1pt,noitemsep]
                \item[\radiobutton*] ``Yes''
                \item[\radiobutton] ``No''
            \end{itemize}
            \item Do the characters  act according to the laws and rules of their time?
            \begin{itemize}[topsep=0.1pt,noitemsep]
                \item[\radiobutton*] ``Yes''
                \item[\radiobutton] ``No''
            \end{itemize}
            \item Do the actants do something that society will approve of?
            \begin{itemize}[topsep=0.1pt,noitemsep]
                \item[\radiobutton*] ``Yes''
                \item[\radiobutton] ``No''
            \end{itemize}
            \item Do the characters  receive a fair retribution/reward/punishment for their actions?
            \begin{itemize}[topsep=0.1pt,noitemsep]
                \item[\radiobutton*] ``Yes''
                \item[\radiobutton] ``No''
            \end{itemize}
            \item Have the people in the text become wealthier and happier without making others much more unhappy?
            \begin{itemize}[topsep=0.1pt,noitemsep]
                \item[\radiobutton*] ``Yes''
                \item[\radiobutton] ``No''
            \end{itemize}
        \end{itemize}

    \noindent \underline{Explanation}: A toddler and the old lady have shown their best spiritual qualities. Both acted according to the law. Society usually approves of such behavior. The good deed was rewarded with justice. Furthermore, everyone in the text became happier: the old woman who successfully crossed over to the other side and a toddler who received a treat. Please answer ``Yes'' to all five questions.

\par\noindent\rule{\textwidth}{0.8pt}

\vspace{0.05cm}
\textbf{Example of web interface}
\vspace{0.1cm}

\colorbox{Gray}{This is a toy text.}

\vspace{0.05cm}
\noindent \textit{Please answer the questions:}

\begin{itemize}[noitemsep,topsep=0.05pt]
            \item Do the characters in this text act with the best intentions, showing their kindest character traits and spiritual qualities?
            \begin{itemize}[topsep=0.1pt,noitemsep]
                \item[\radiobutton] ``Yes''
                \item[\radiobutton] ``No''
            \end{itemize}
            \item Do the characters  act according to the laws and rules of their time?
            \begin{itemize}[topsep=0.1pt,noitemsep]
                \item[\radiobutton] ``Yes''
                \item[\radiobutton] ``No''
            \end{itemize}
            \item Do the actants do something that society will approve of?
            \begin{itemize}[topsep=0.1pt,noitemsep]
                \item[\radiobutton] ``Yes''
                \item[\radiobutton] ``No''
            \end{itemize}
            \item Do the characters  receive a fair retribution/reward/punishment for their actions?
            \begin{itemize}[topsep=0.1pt,noitemsep]
                \item[\radiobutton] ``Yes''
                \item[\radiobutton] ``No''
            \end{itemize}
            \item Have the people in the text become wealthier and happier without making others much more unhappy?
            \begin{itemize}[topsep=0.1pt,noitemsep]
                \item[\radiobutton] ``Yes''
                \item[\radiobutton] ``No''
            \end{itemize}
        \end{itemize}
\par\noindent\rule{\textwidth}{1pt}

\caption{The instruction for the \textbf{Ethics$_2$} human evaluation project translated for illustration purposes.}

\label{tab:ethics2}
\end{minipage}

\end{table*}

\clearpage
\newpage

\onecolumn
\section{Diagnostic Analysis}
\label{app:subpopulations}
\begin{figure*}[htb!]
    \centering
    \begin{subfigure}[b]{0.72\textwidth}
        \includegraphics[width=1\linewidth]{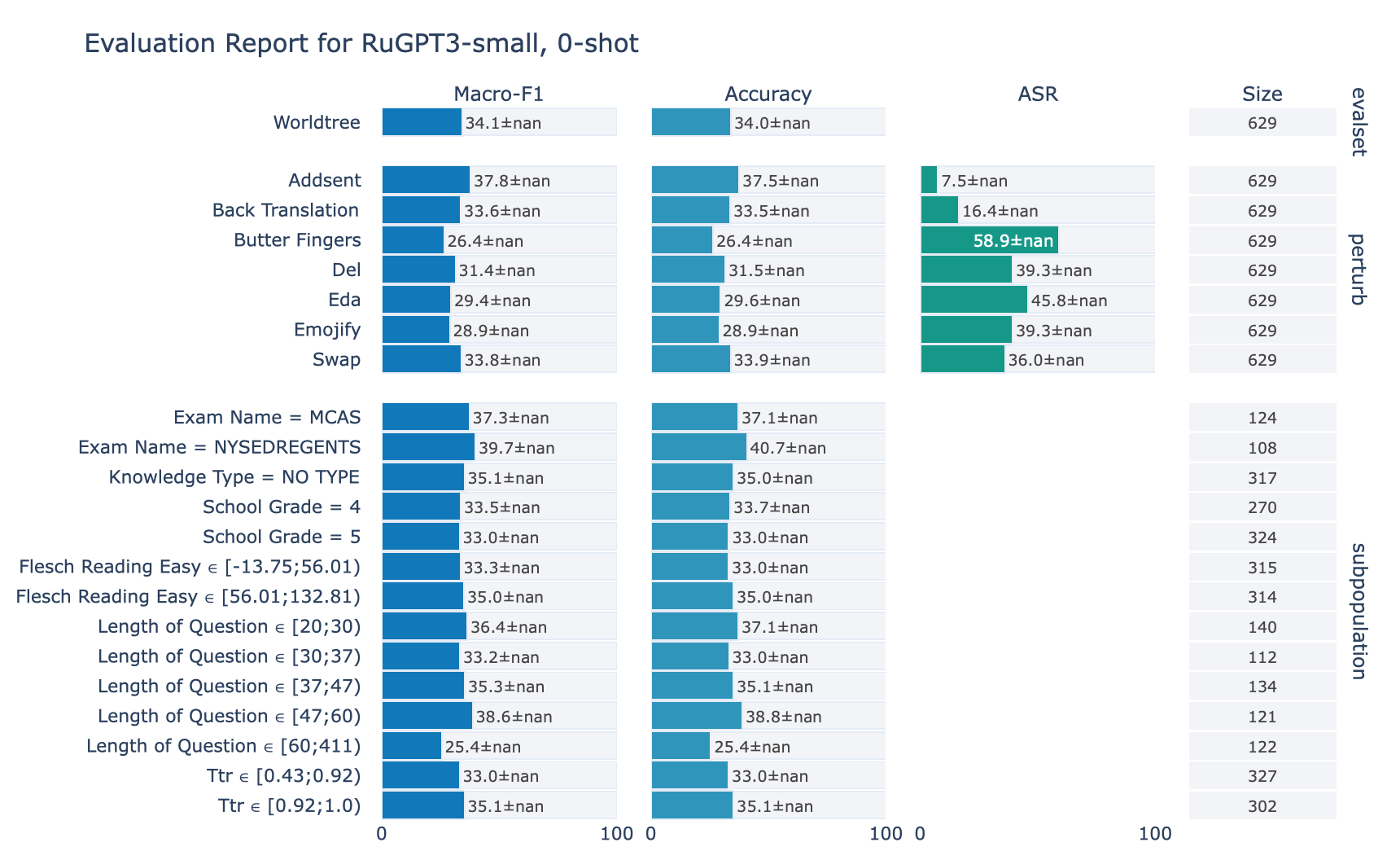}
        \caption{ruGPT$_\textsc{S}$}
    \end{subfigure}
    \begin{subfigure}[b]{0.72\textwidth}
        \includegraphics[width=1\linewidth]{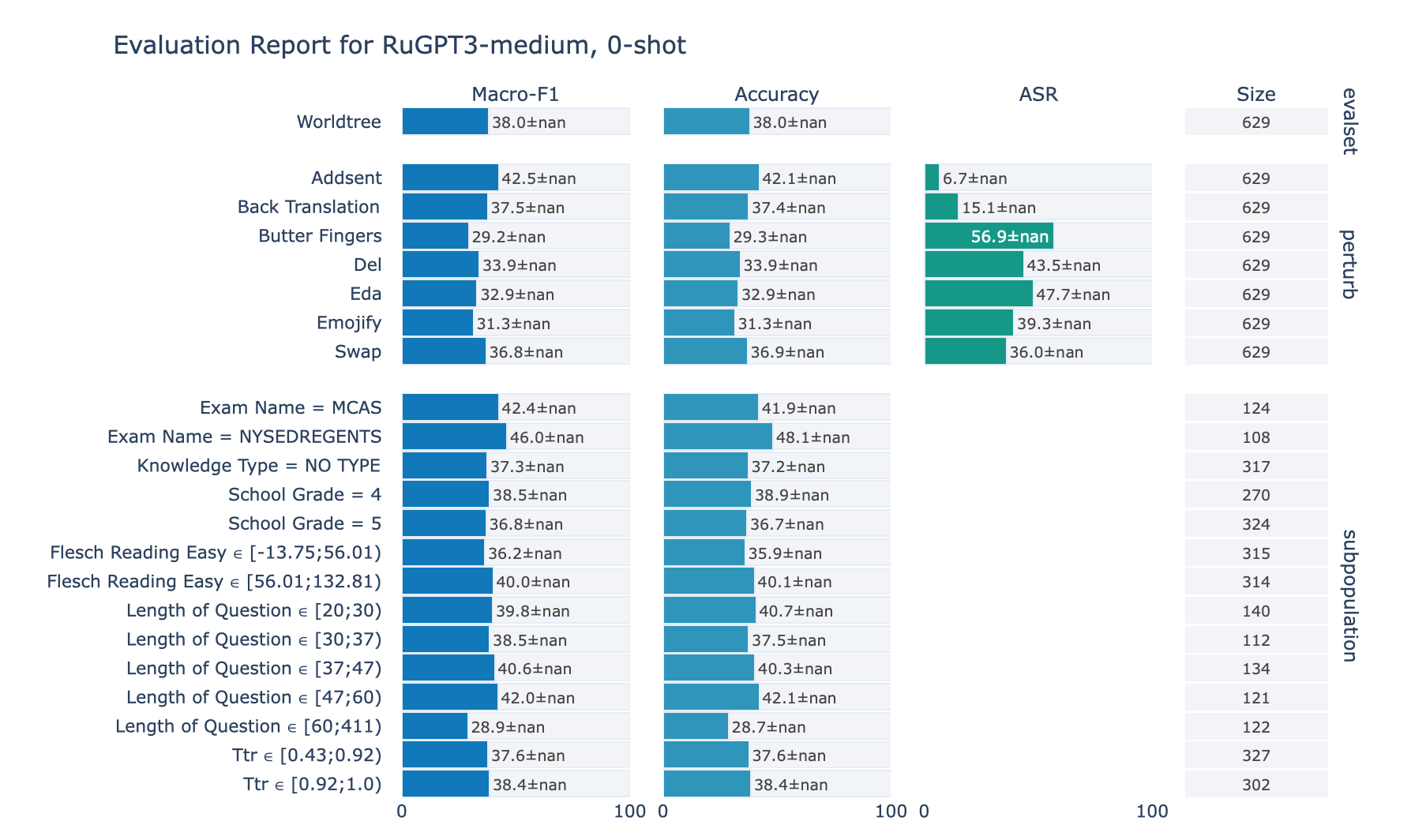}
        \caption{ruGPT$_\textsc{M}$}
    \end{subfigure}
    \begin{subfigure}[b]{0.72\textwidth}
        \includegraphics[width=1\linewidth]{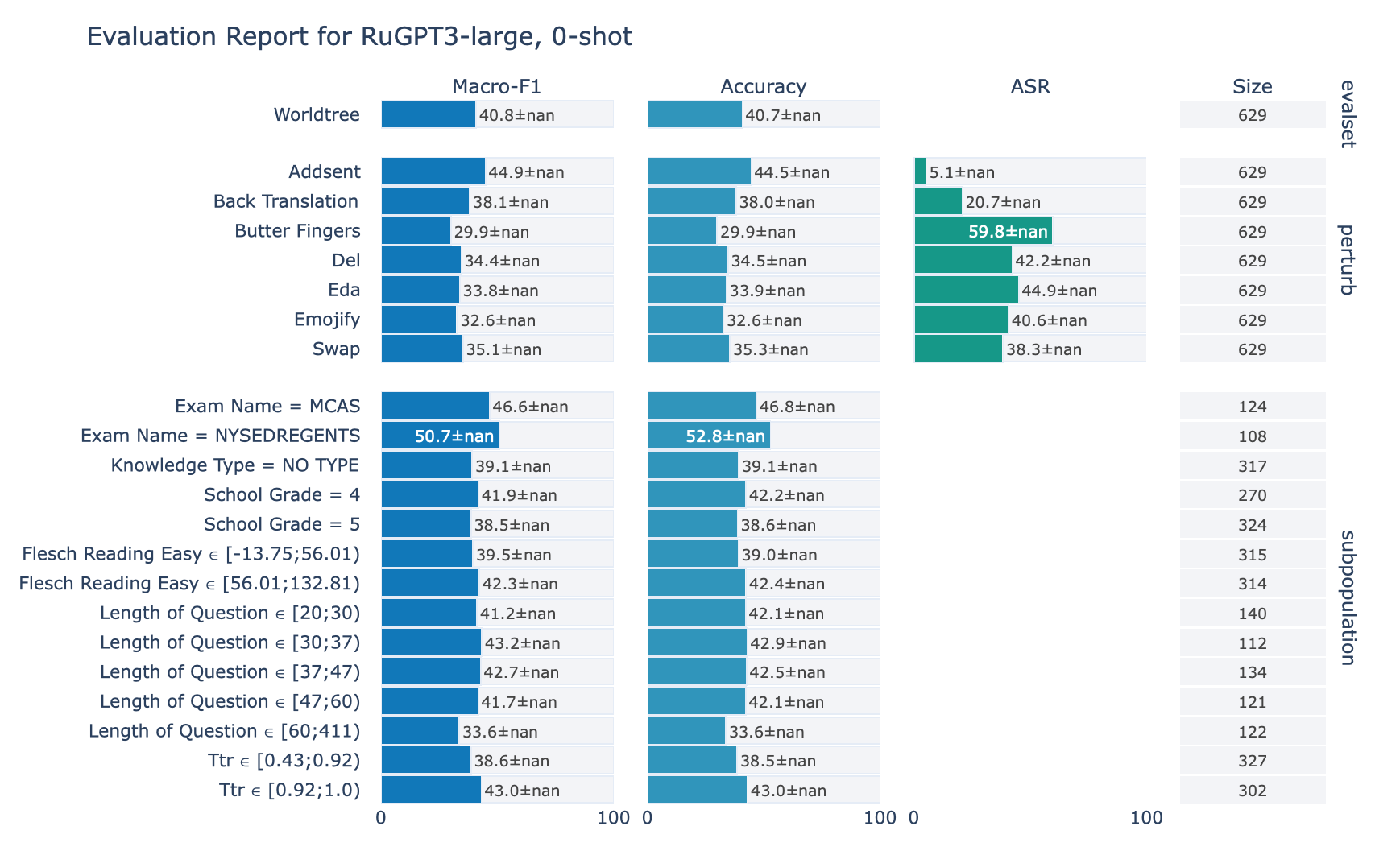}
        \caption{ruGPT$_\textsc{L}$}
    \end{subfigure}
    \caption{Evaluation report for ruGPT models on the \textbf{RuWorldTree} task in the $0$-shot setting.}
    \label{fig:report0}
\end{figure*}

\begin{figure*}[p!]
    \centering
    \begin{subfigure}[b]{0.75\textwidth}
        \includegraphics[width=1\linewidth]{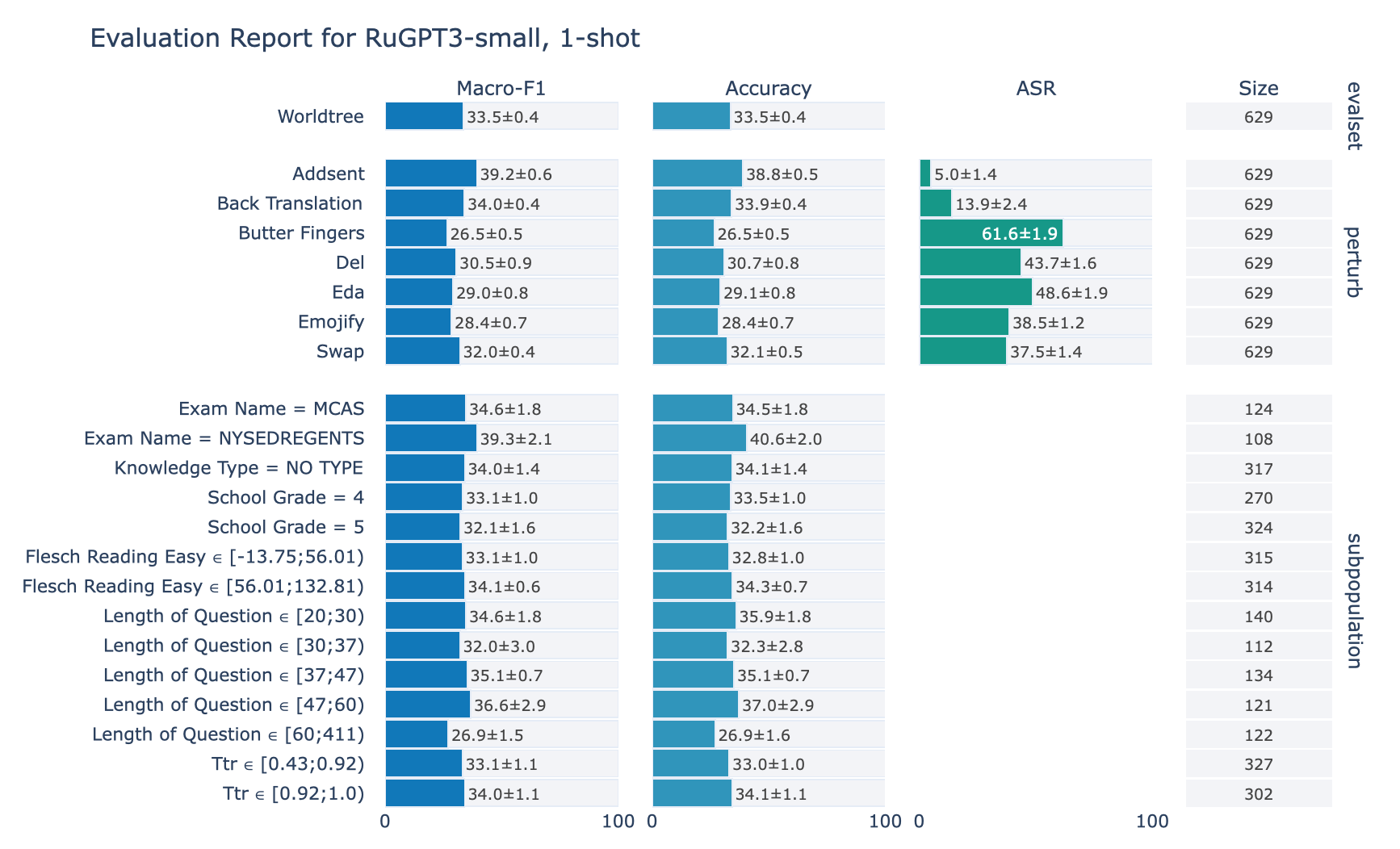}
        \caption{ruGPT$_\textsc{S}$}
    \end{subfigure}
    \begin{subfigure}[b]{0.75\textwidth}
        \includegraphics[width=1\linewidth]{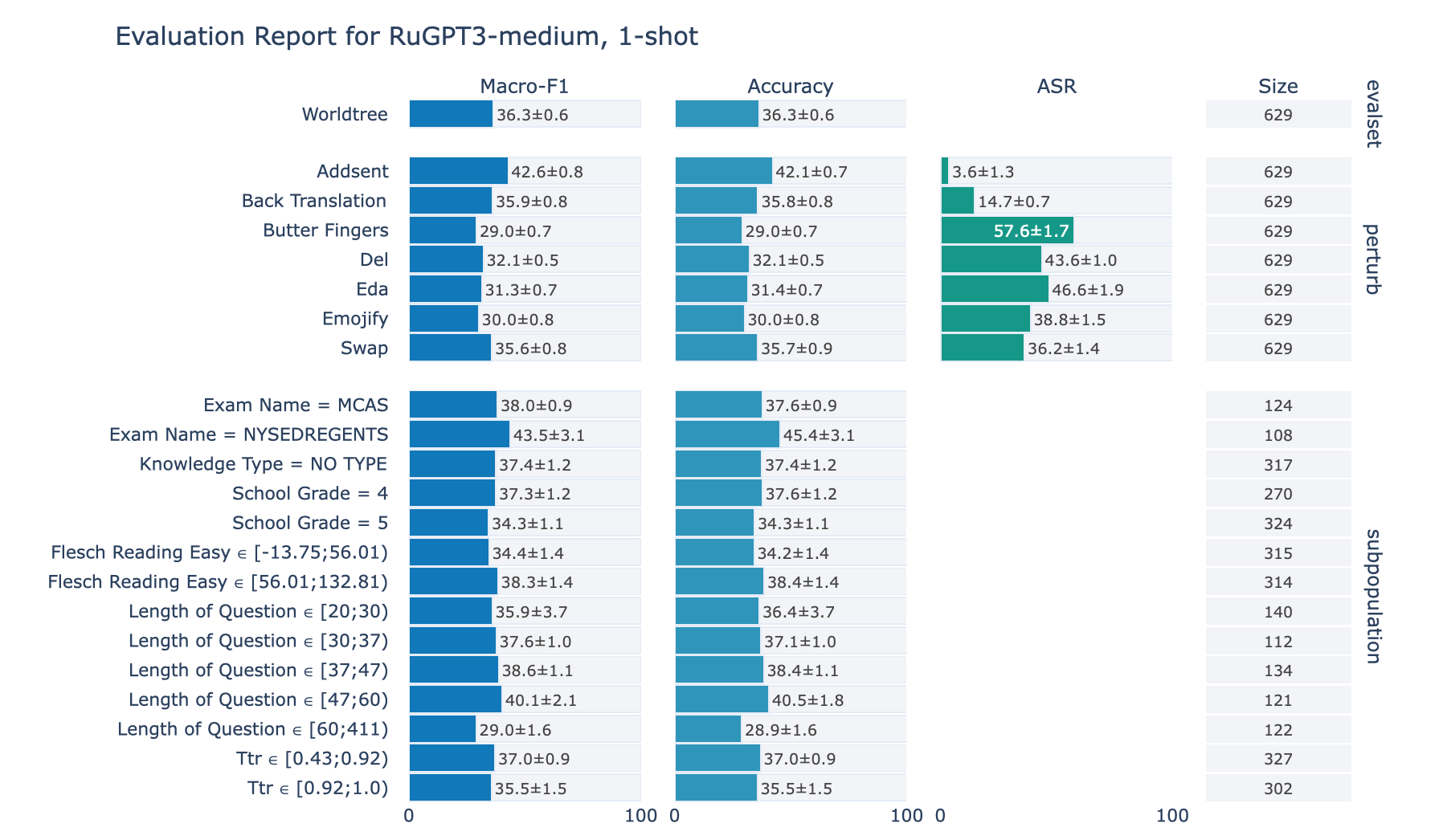}
        \caption{ruGPT$_\textsc{M}$}
    \end{subfigure}
    \begin{subfigure}[b]{0.75\textwidth}
        \includegraphics[width=1\linewidth]{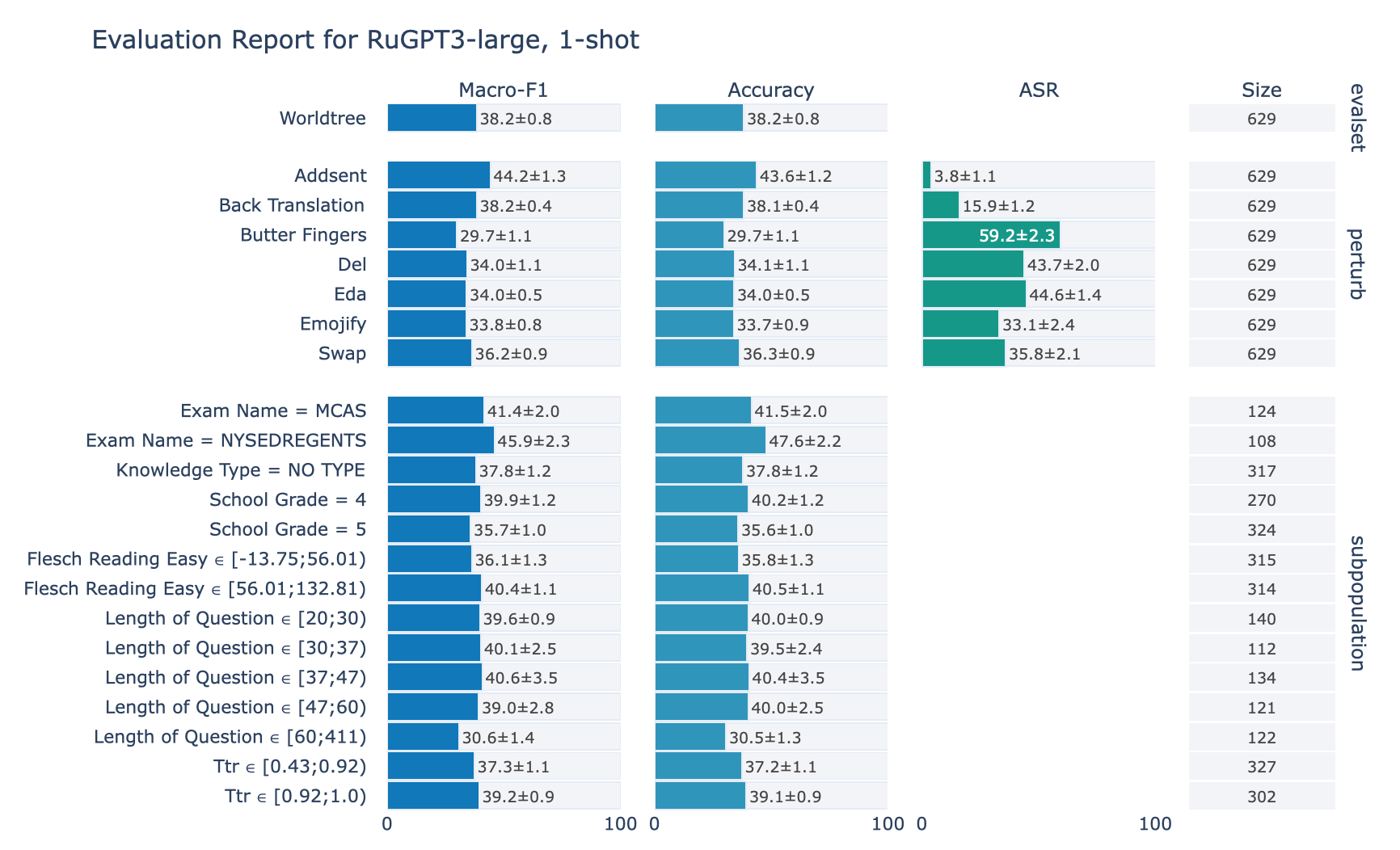}
        \caption{ruGPT$_\textsc{L}$}
    \end{subfigure}
    \caption{Evaluation report for ruGPT models on the \textbf{RuWorldTree} task in the $1$-shot setting.}
    \label{fig:report1}
\end{figure*}

\begin{figure*}[p!]
    \centering
    \begin{subfigure}[b]{0.75\textwidth}
        \includegraphics[width=1\linewidth]{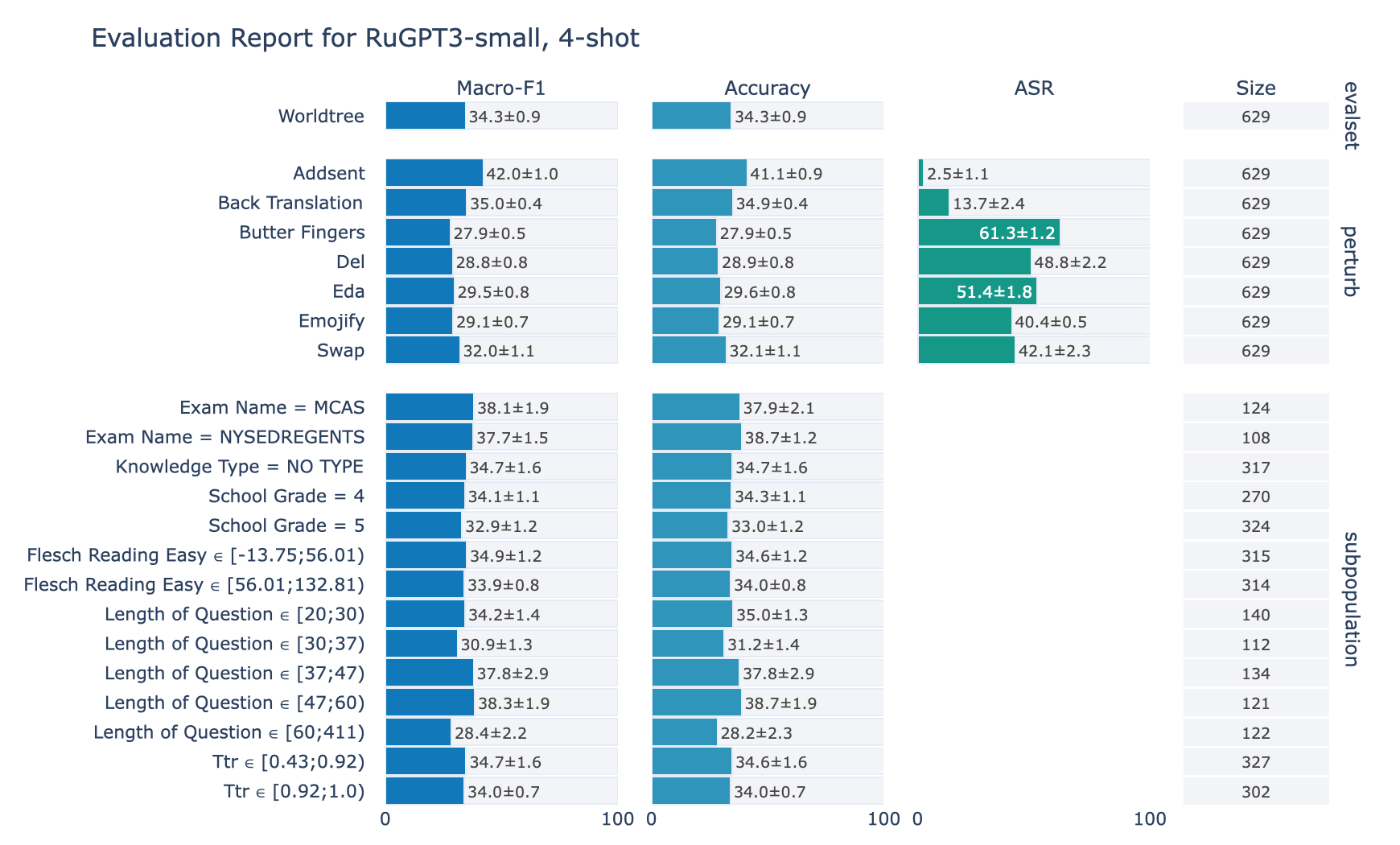}
        \caption{ruGPT$_\textsc{S}$}
    \end{subfigure}
    \begin{subfigure}[b]{0.75\textwidth}
        \includegraphics[width=1\linewidth]{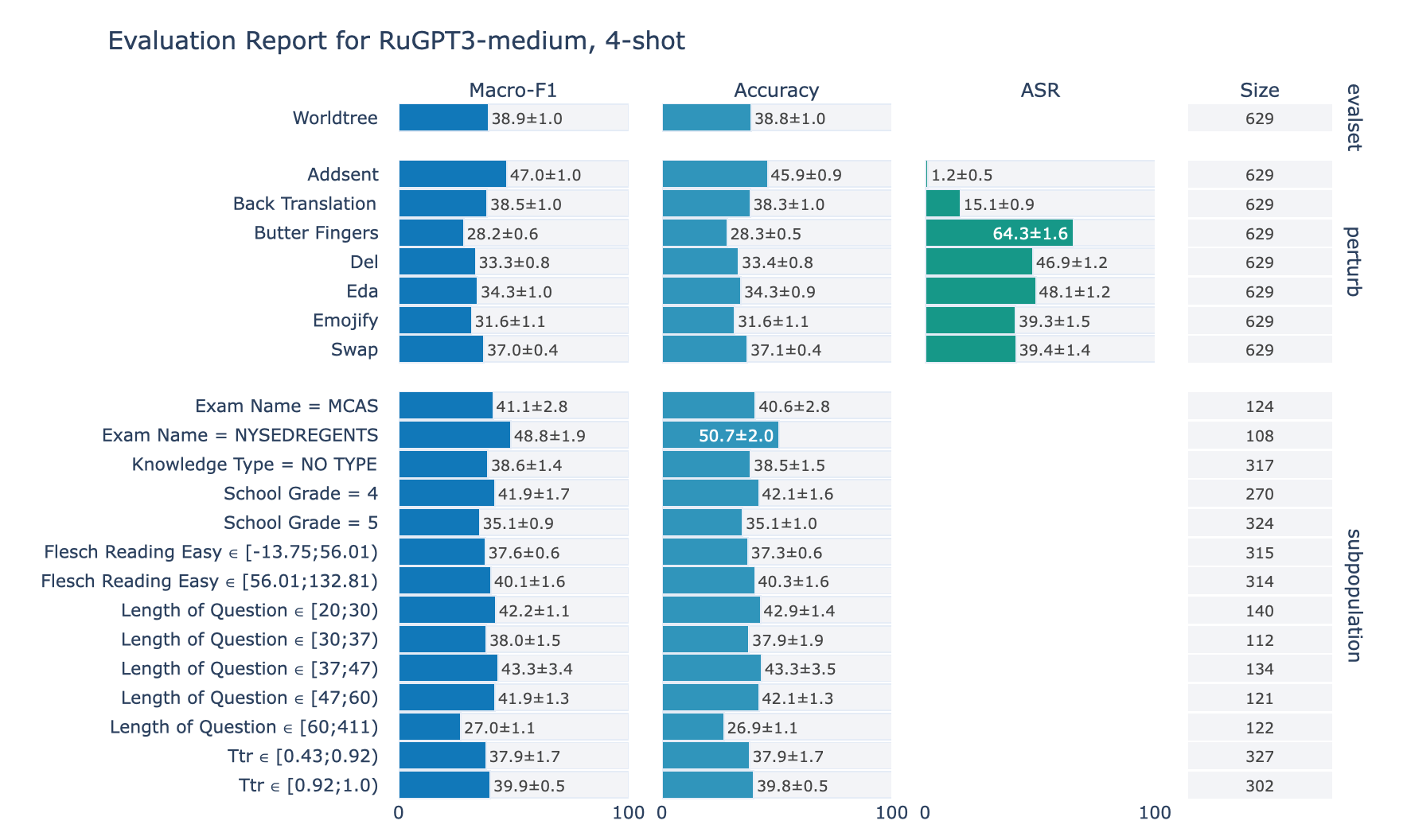}
        \caption{ruGPT$_\textsc{M}$}
    \end{subfigure}
    \begin{subfigure}[b]{0.75\textwidth}
        \includegraphics[width=1\linewidth]{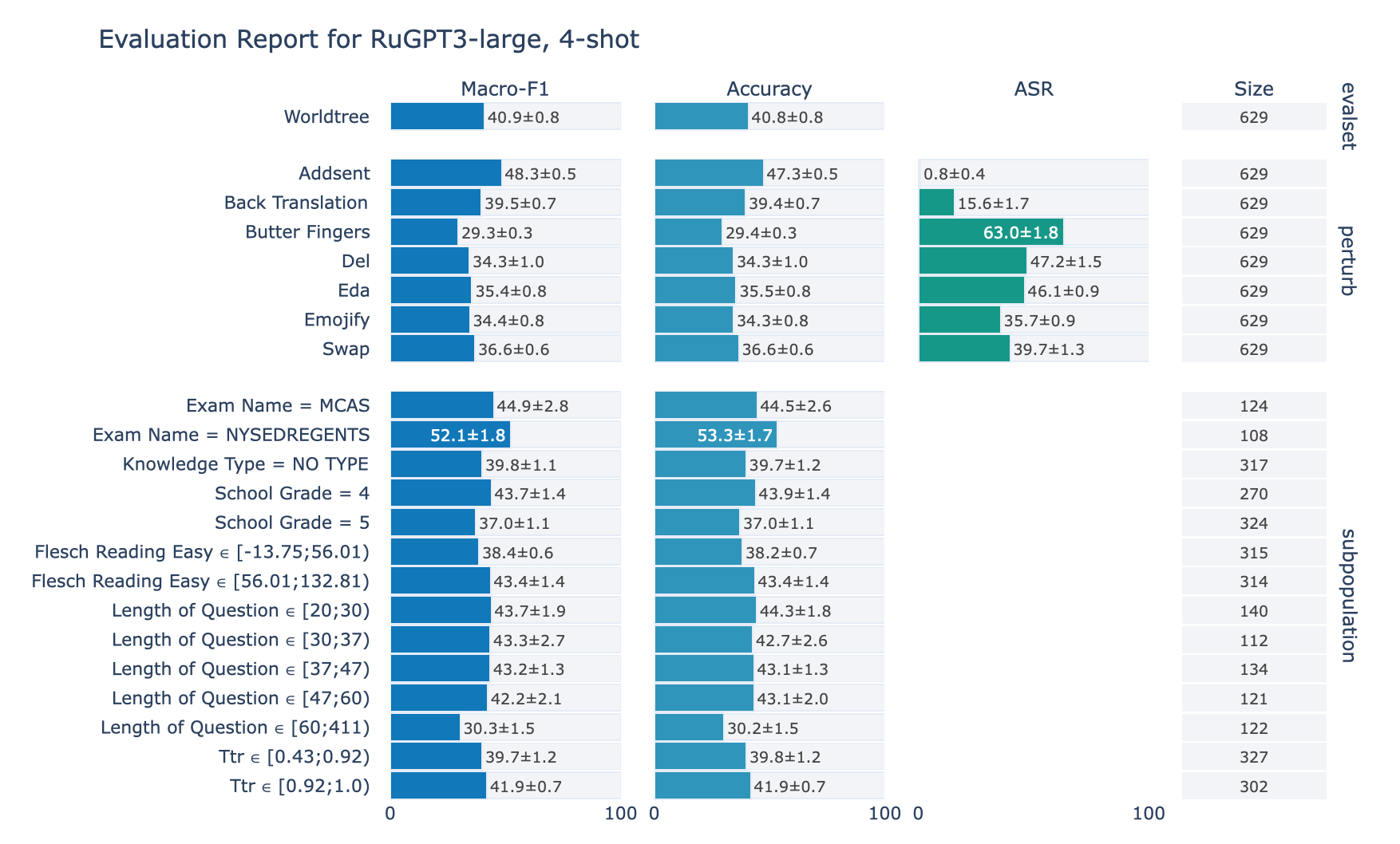}
        \caption{ruGPT$_\textsc{L}$}
    \end{subfigure}
    \caption{Evaluation report for ruGPT models on the \textbf{RuWorldTree} task in the $4$-shot setting.}
    \label{fig:report4}
\end{figure*}

\begin{figure*}[p!]
    \centering
    \begin{subfigure}[b]{0.75\textwidth}
        \includegraphics[width=1\linewidth]{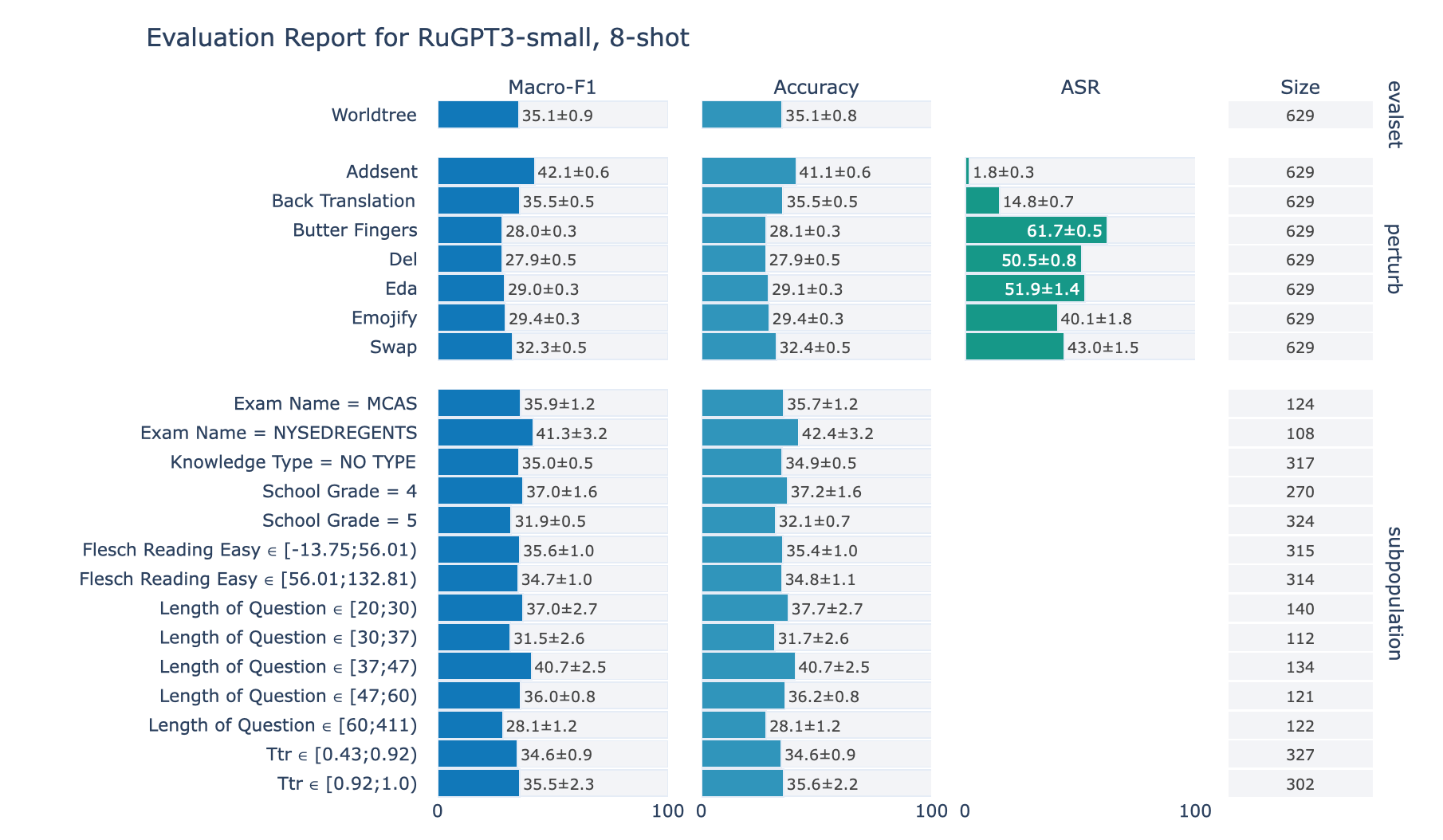}
        \caption{ruGPT$_\textsc{S}$}
    \end{subfigure}
    \begin{subfigure}[b]{0.75\textwidth}
        \includegraphics[width=1\linewidth]{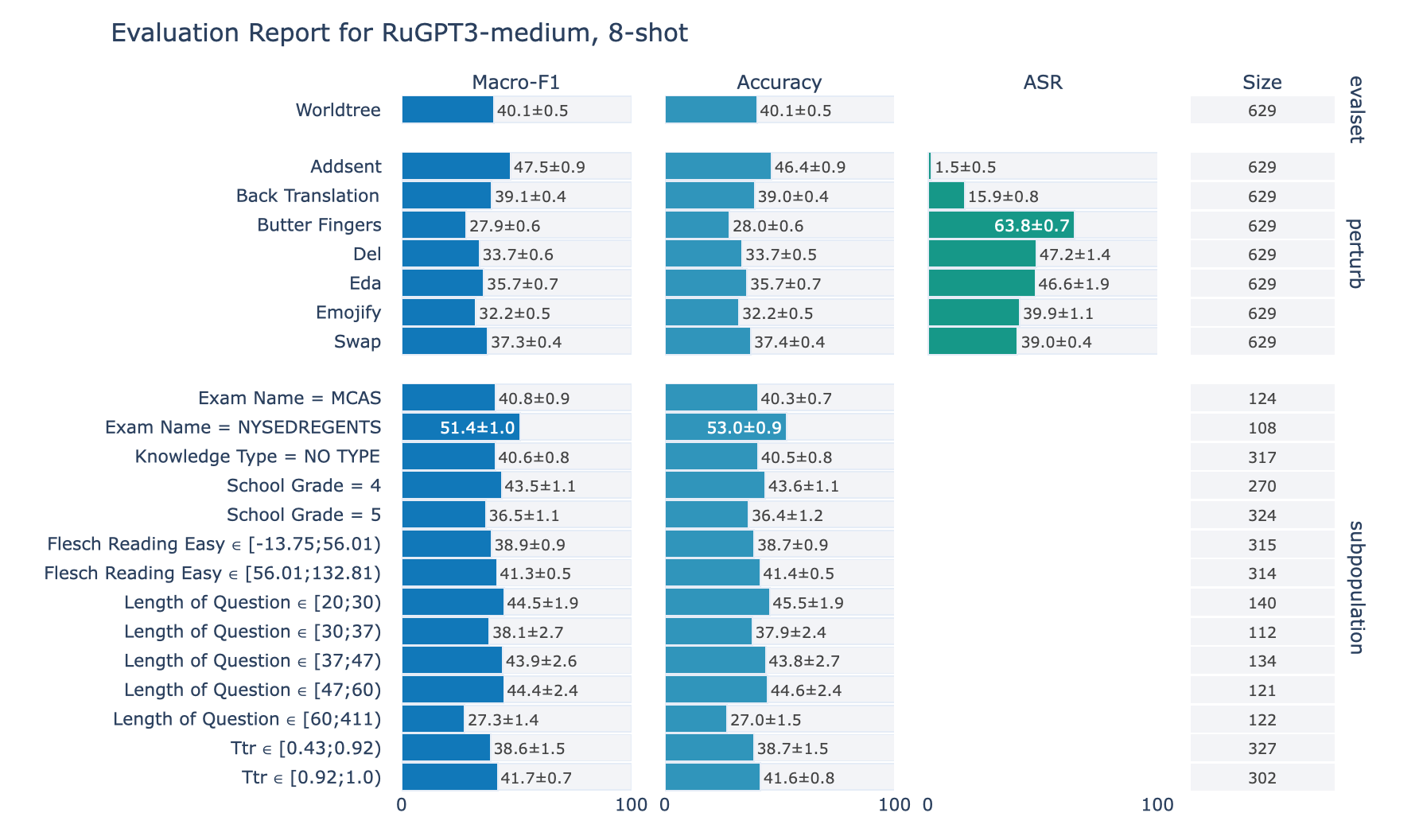}
        \caption{ruGPT$_\textsc{M}$}
    \end{subfigure}
    \begin{subfigure}[b]{0.75\textwidth}
        \includegraphics[width=1\linewidth]{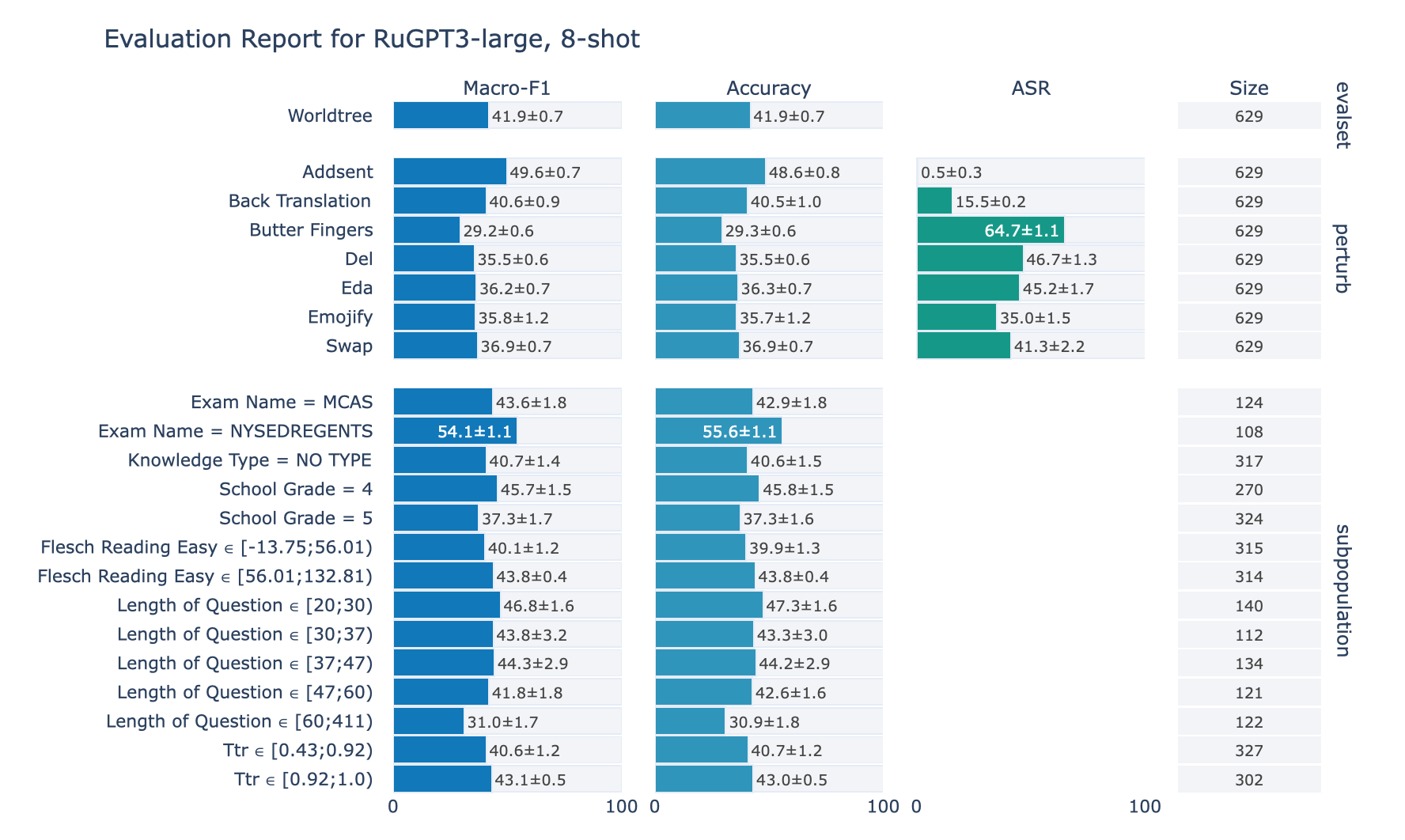}
        \caption{ruGPT$_\textsc{L}$}
    \end{subfigure}
    \caption{Evaluation report for ruGPT models on the \textbf{RuWorldTree} task in the $8$-shot setting.}
    \label{fig:report8}
\end{figure*}

\end{document}